%% file: main.tex
\documentclass[twoside]{article}

\usepackage[accepted]{aistats2026}
\usepackage[round]{natbib}

\usepackage{graphicx}

\usepackage{caption}
\usepackage{subcaption} 
\usepackage{hyperref}

\newcommand{\subindex}{{ j }}
\newcommand{\modalities}{{ M }}

\usepackage{xcolor}         
\usepackage{multirow}
\usepackage{graphicx}
\def\rot{\rotatebox}

\usepackage{titletoc}
\usepackage[page,header]{appendix}
\usepackage{booktabs}
\usepackage{tablefootnote} 
\usepackage{makecell}

\input{math_commands.tex}

\begin{document}

%

%

\twocolumn[

\aistatstitle{Hellinger Multimodal Variational Autoencoders}

\aistatsauthor{ Huyen Vo $^{\dagger, \ddagger}$ \And Isabel Valera $^{\dagger, \ddagger}$}

\aistatsaddress{ $^{\dagger}$ Department of Computer Science, Saarland University, DE \\ $^{\ddagger}$ MPI-SWS, Saarland Informatics Campus, DE} ]

\begin{abstract}
  Multimodal variational autoencoders (VAEs) are widely used for weakly supervised generative learning with multiple modalities. Predominant methods aggregate unimodal inference distributions using either a product of experts (PoE), a mixture of experts (MoE), or their combinations to approximate the joint posterior. In this work, we revisit multimodal inference through the lens of probabilistic opinion pooling, an optimization-based approach. We start from H\"older pooling with $\alpha=0.5$, which corresponds to the unique symmetric member of the $\alpha\text{-divergence}$ family, and derive a moment-matching approximation, termed Hellinger. We then leverage such an approximation to propose HELVAE, a multimodal VAE that avoids sub-sampling, yielding an efficient yet effective model that: (i) learns more expressive latent representations as additional modalities are observed; and (ii) empirically achieves better trade-offs between generative coherence and quality, outperforming state-of-the-art multimodal VAE models. 
\end{abstract}


\section{INTRODUCTION}

Recently, frameworks for cross-generation, such as image-to-text~\citep{li2021align} and text-to-image~\citep{radford2021learning, nichol2021glide}, have received attention~\citep{li2019controllable, zhang2021cross, nichol2021glide, saharia2022photorealistic, gu2022vector}. However, these advances focus on cross-modality learning through conditional dependencies from one modality to another, often neglecting shared semantic representations across modalities. In practice, multimodal datasets are weakly supervised, as collecting complete annotations is costly, and access to all modalities at inference time is rarely guaranteed. Multimodal VAEs address this by providing a probabilistic framework that learns joint latent representations capturing shared and modality-specific structures, while also enabling cross-generation, missing-modality inference, and weakly supervised learning.

To handle missing-modality inference, early multimodal VAEs~\citep{suzuki2016joint, vedantam2017generative} lacked scalability, as they required a separate encoder for each subset of modalities. More recent methods~\citep{wu2018multimodal, shi2019variational, sutter2021generalized} introduce scalable inference mechanisms that efficiently learn from many modalities using a joint encoder factorized into unimodal components. These approaches primarily follow two paradigms: product of experts (PoE), such as MVAE~\citep{wu2018multimodal}, and mixture of experts (MoE), such as MMVAE~\citep{shi2019variational}. Both, however, have flaws in estimating the joint posterior: PoE can be biased when expert precisions are miscalibrated or collapse if any expert assigns low density, while MoE cannot yield a posterior sharper than its components, limiting efficiency in high-dimensional spaces~\citep{hinton2002training}. Their performance is typically assessed through two criteria: generative quality, which measures how well the model fits the data distribution, and generative coherence, which evaluates semantic consistency across modalities. Thus, an effective multimodal generative model should ideally achieve both. However, in~\citet{daunhawer2021limitations}, the authors show that existing approaches struggle to balance these objectives, limiting applicability to complex real-world settings. In particular, they note that mixture-based models such as MMVAE rely on sub-sampling,\footnote{Sub-sampling refers to training multimodal VAEs by randomly selecting subsets of modalities instead of using all modalities at once.} which imposes an undesirable upper bound on the multimodal ELBO and limits generative quality~\citep{daunhawer2021limitations}.


In this work, we study PoE and MoE aggregation through probabilistic opinion pooling, and more specifically,  H\"older pooling, which corresponds to minimizing $\alpha$-divergence~\citep{amari1985differential, zhu1995information}. In this perspective, PoE can be seen as log-linear pooling~\citep{genest1984characterization} when $\alpha \to 0$ and MoE as linear pooling~\citep{stone1961opinion} when $\alpha = 1$. We then leverage the unique symmetric case of the $\alpha$-divergence family, $\alpha=0.5$, proportional to the squared Hellinger distance.  This distance is a bounded metric on probability measures, defined as the $\normltwo$ norm between square-root densities~\citep{nikulin2001hellinger}. Building on this, we apply moment matching to project the pooled distribution onto a single Gaussian, allowing multimodal VAEs avoid sub-sampling. We then leverage this approximation to propose the \textbf{HEL}linger multimodal \textbf{VAE} (HELVAE), an efficient yet effective model without sub-sampling during training.

In summary, our main contributions are three-fold: (i) We introduce a probabilistic opinion pooling perspective for multimodal VAEs, which generalizes common methods such as PoE and MoE as special cases of H\"older pooling with respective $\alpha$; (ii) we develop a moment-matching approximation of H\"older pooling with $\alpha=0.5$, termed Hellinger aggregation, which is particularly effective when combining different types of experts; and,  building on this, (iii) we then propose HELVAE, a multimodal VAE that provides an efficient multimodal VAE that balances generative coherence and quality. 
Our empirical results on three benchmark datasets (PolyMNIST, CUB Image-Captions, and bimodal CelebA), show that HELVAE learns better latent representations and shows synergy~\footnote{Synergy~\citep{shi2019variational} is the effect that learning from multiple modalities improves generative performance in each modality.} across modalities, leading to  better trade-offs between generative coherence and quality than existing  methods.


\section{RELATED WORK}

Extending VAEs~\citep{kingma2013auto} to multiple data modalities, early multimodal VAEs~\citep{suzuki2016joint, vedantam2017generative} lacked scalability. More recent approaches address this using PoE~\citep{wu2018multimodal}, MoE~\citep{shi2019variational}, or their combination, the mixture of product of experts (MoPoE)~\citep{sutter2021generalized}. Building on this line, later work has introduced WBVAE~\citep{qiu2025multimodal}, which leverages Wasserstein barycenter aggregation, and CoDEVAE~\citep{mancisidor2025aggregation}, which defines joint distributions through a mixture of consensus among dependent experts (CoDE). Although CoDEVAE claims it does not require sub-sampling, since all loss terms for all subsets are trained rather than some as in mixture-based methods, CoDEVAE still incurs a relatively high computational cost of $\mathcal{O}(2^M - 1)$, similar to MoPoE and MWBVAE, where $M$ is the number of modalities. Beyond these approaches, flexible aggregation methods based on permutation-invariant neural networks~\citep{hirtlearning} have also been proposed.

In this context,~\citet{daunhawer2021limitations} show that multimodal VAEs struggle with a trade-off between generative coherence and quality. ~\citet{javaloy2022mitigating} further identify modality collapse as a consequence of conflicting gradients during training. Since then, performance has been improved by regularization objectives such as mmJSD~\citep{sutter2020multimodal}, which aligns unimodal and joint posteriors via Jensen–Shannon divergence, and MVTCAE~\citep{hwang2021multi}, which captures cross-view dependencies via total correlation. In addition, hierarchical latent spaces have been leveraged to capture heterogeneity among modalities~\citep{vasco2022leveraging, wolffhierarchical}. Another line of work has focused on dynamic priors and posteriors, using unimodal posteriors as priors~\citep{sutter2020multimodal, joy2021learning}, energy-based priors~\citep{yuan2024learning}, variational mixtures of posteriors (Vamp) priors~\citep{sutter2024unity}, score-based priors~\citep{wesego2024scorebased}, Markov random field~\citep{oubari2024markov}, and flow-based posteriors~\citep{senellart2025bridging}. More recently, diffusion models have been used in multimodal VAEs, either at the decoder or the latent level~\citep{wesego2024revising, bounoua2024multi, palumbo2024deep}. Finally, prior work has considered separating modality-specific latent subspaces alongside a shared subspace~\citep{sutter2020multimodal, lee2021private, palumbo2023mmvae+}, with MMVAE+~\citep{palumbo2023mmvae+} as the state-of-the-art model that introduces a new objective for mixture-based models.
 
While these approaches consider different aspects of multimodal representation learning, they generally use the two most common combination schemes and are orthogonal to this work, which specifically focuses on the modality aggregation approach. We compare our model with recent aggregation-based methods, as well as MMVAE+ with additional modality-specific latent variables, and show that it achieves comparable generative performance without requiring extra complexity.


\section{PRELIMINARIES}

\subsection{Multimodal VAEs}

We consider multimodal data $\sX$, comprising $\modalities$ modalities $ \vx_1, \vx_2, \dots, \vx_\modalities$, which are assumed to be generated from a shared latent variable $ \vz$ using modality-specific parametric decoders $ \{p_{\theta_j}(\vx_\subindex|\vz)\}_{\subindex=1}^\modalities$. Assuming conditional independence between modalities given $ \vz$, the joint distribution factorizes as: $ \textstyle p_\theta(\sX,\vz)  = p(\vz)\prod_{\subindex=1}^\modalities p_{\theta_j}(\vx_\subindex|\vz)$, where $ p(\vz)$ is the prior over the latent space. The objective of a multimodal VAE is to maximize the log-likelihood of data over $\modalities$ modalities: 
\begin{equation*}
    \log p_\theta(\sX) = \KL(q_\phi(\vz|\sX) \Vert p_\theta(\vz|\sX)) + \mathcal{L}(\theta, \phi; \sX),
\end{equation*}
\begin{equation*}
    \mathcal{L}(\theta, \phi;\sX) = \mathbb{E}_{q_\phi(\vz|\sX)}[\log p_\theta(\sX|\vz)] - \KL(q_\phi(\vz|\sX) \Vert p(\vz)).
\label{eqn:elbo}
\end{equation*}
The distribution $ q_\phi(\vz|\sX)$ is a variational posterior approximation with learnable parameters $\phi$, since the true posterior is intractable. From the non-negativity of the KL divergence, yielding $ \log p_\theta(\sX) \geq \mathcal{L}(\theta, \phi; \sX)$. The term $ \mathcal{L}(\theta, \phi; \sX)$ is the evidence lower bound (ELBO) on the marginal log-likelihood, and we maximize this bound instead. Maximizing the ELBO requires approximating the joint posterior $q_\phi(\vz|\sX)$ in a way that scales flexibly to many modalities. Two common approaches are PoE~\citep{wu2018multimodal} and MoE~\citep{shi2019variational}, where the joint distributions are approximated as $q_\phi(\vz|\sX) = c\prod_{\subindex=1}^\modalities q_{\phi_\subindex}(\vz|\vx_\subindex)$ and $q_\phi(\vz|\sX) = \frac{1}{\modalities} \sum_{\subindex=1}^\modalities q_{\phi_\subindex}(\vz|\vx_\subindex)$, respectively, with $c$ a normalization constant that guarantees the validity of the resulting probability measure.

\subsection{Probabilistic opinion pooling: an optimization-based approach}

Probabilistic opinion pooling is a principled framework for aggregating multiple probability density functions (pdfs) into a single consensus pdf. Let $\vz = (\vz_1, \vz_2, \dots, \vz_D)^\top \in \sR^D$ be a continuous random variable. Given a set of pdfs $ \{q_j(\vz)\}_{j=1}^M \in \mathcal{P}^M$ with associated non-negative weights $\{\lambda_j\}_{j=1}^M$ satisfying $\sum_{\subindex=1}^\modalities \lambda_\subindex = 1$, a pooled density is defined as a weighted aggregation of these pdfs. A popular pooling function is linear pooling~\citep{stone1961opinion}, which computes a weighted arithmetic average: $ \textstyle q(\vz) = \sum_{\subindex=1}^\modalities \lambda_\subindex q_\subindex(\vz)$. Another popular one is the log-linear pooling function~\citep{genest1984characterization}, which takes a weighted geometric average: $ \textstyle q(\vz) = c \prod_{\subindex=1}^\modalities (q_\subindex(\vz))^{\lambda_\subindex}$ where c is a normalization factor. 

In an optimization-based approach, the pooling function is obtained by minimizing a weighted average of some discrepancy measure between the aggregated pdf and the individual pdfs. One class of discrepancy measures that we consider in this paper is the family of $f$-divergences. Given a convex function $ f: \sR^+ \rightarrow \sR$ satisfying $f(1)=0$, the $f$-divergence between two pdfs $ q_\subindex(\vz)$ and $ \varphi(\vz)$ over the domain $\sR^D$ is defined as:
\begin{equation*}
 \mathcal{D}_f(q_\subindex \Vert \varphi) = \int \varphi(\vz) f\left( \frac{q_\subindex(\vz)}{\varphi(\vz)} \right) d\vz.  
\end{equation*}
The aggregated pdf $ q(\vz)$ is obtained by solving the following minimization problem:
\begin{equation}
 q(\vz) = \argmin_{\varphi \in \mathcal{P}} \sum_{\subindex=1}^\modalities \lambda_\subindex \mathcal{D}_f(q_\subindex \Vert \varphi).
\label{eqn:min_fdiv}
\end{equation}
As stated in~\citet{Abbas2009AKV, qiu2025multimodal}, log-linear pooling and linear pooling can be derived as solutions to the minimization problem in Equation~\ref{eqn:min_fdiv}, corresponding respectively to the reverse and forward KL divergences, which are both members of the $f$-divergence family: $f(x) = -\log x$ for the reverse KL and $f(x) = x \log x$ for the forward KL. We generalize these two special cases to an entire family of divergences and corresponding pooling functions, both parameterized by a real-valued parameter $\alpha$. Specifically, we consider $f(x) = \frac{x^\alpha - 1}{\alpha(\alpha - 1)}$ for $x > 0$ and $ \alpha \in \sR \backslash \{0, 1\}$. This yields the family of $\alpha$-divergences defined as:
\begin{equation*}
 \mathcal{D}_{\alpha}(q_\subindex \Vert \varphi) = \frac{1}{\alpha(\alpha - 1)} \int \varphi(\vz)  \frac{(q_\subindex(\vz))^\alpha - (\varphi(\vz))^\alpha}{(\varphi(\vz))^\alpha} d\vz.
\label{eqn:alphadiv}
\end{equation*}
The solution to the problem in Equation~\ref{eqn:min_fdiv} for the $\alpha$-divergences is an $\alpha$-parameterized family of H\"older pooling functions, as mentioned in~\citet{Garg2004GeneralizedOP, koliander2022fusion}:
\begin{equation*}
 q(\vz) = c \left( \sum_{\subindex=1}^\modalities \lambda_\subindex (q_\subindex(\vz))^\alpha \right)^{1/\alpha},
\end{equation*}
where $ \textstyle c = 1/ \int \left( \sum_{\subindex=1}^\modalities \lambda_\subindex (q_\subindex(\vz))^\alpha \right)^{1/\alpha} d\vz$. In the limit $\alpha \to 0$, the H\"older pooling function reduces to the log-linear pooling function, and for $\alpha = 1$, it becomes the linear pooling function, consistent with the facts that $\textstyle \lim_{\alpha \to 0} \mathcal{D}_{\alpha}(q_\subindex \Vert \varphi) = \KL(\varphi \Vert q_\subindex)$ (reverse KL) and $\textstyle \lim_{\alpha \to 1} \mathcal{D}_{\alpha}(q_\subindex \Vert \varphi) = \KL(q_\subindex \Vert \varphi)$ (forward KL)~\citep{minka2005divergence}. It is worth noting that the above choices of $\alpha$ yield asymmetric and unbounded discrepancy measures, which do not define a metric space for probability measures. In what follows, we consider the unique symmetric case of the $\alpha$-divergence family, corresponding to $\alpha = 0.5$:
\begin{align*}
 \mathcal{D}_{0.5}(q_\subindex \| \varphi) &= 2 \int \left( \sqrt{q_\subindex(\vz)} - \sqrt{\varphi(\vz)} \right)^2 d\vz \\ &= 4 \text{Hel}^2(q_\subindex \Vert \varphi),
\end{align*}
where $\text{Hel}(q_\subindex \Vert \varphi)$ is the Hellinger distance~\citep{nikulin2001hellinger} between two distributions, symmetric and bounded between $0$ and $1$~\citep{tsybakov2009introduction}.

The use of $\alpha$-divergence with $\alpha = 0.5$ was introduced in Black-Box $\alpha$ (BB-$\alpha$)~\citep{hernandez2016black}, an approximate inference method based on the minimization of $\alpha$-divergence. Their experiments demonstrate that BB-$\alpha$ with non-standard settings of $\alpha$, such as $\alpha=0.5$, often yields better predictive performance compared to $\alpha \to 0$ variational Bayes (VB) or $\alpha=1$ expectation propagation (EP). This motivates our use of $\alpha$-divergence with $\alpha = 0.5$ in H\"older pooling, which is defined as:
\begin{align}
q(\vz) &= c \left( \sum_{\subindex=1}^\modalities \lambda_\subindex \sqrt{q_\subindex(\vz)} \right)^2 \label{eqn:holder} \\
&= c \left( \sum_{j=1}^M \lambda_j^2 q_j(\vz) + 2 \sum_{i=1}^M \sum_{j>i}^M \lambda_i \lambda_j \sqrt{q_i(\vz)q_j(\vz)} \right), \nonumber
\end{align}
where $ \textstyle c = 1/ \int \left( \sum_{\subindex=1}^\modalities \lambda_\subindex \sqrt{q_\subindex(\vz)} \right)^{2} d\vz$. 

We note that $q(\vz)$ takes the form of a squared mixture model, closely related to recent squared subtractive mixture formulations, which have been shown to provide increased expressiveness for density estimation tasks~\citep{loconte2023subtractive, loconte2025sum, loconte2025square}. This suggests that exploring negative pooling weights could be a promising direction for future work.


\section{METHODOLOGY}

By applying the probabilistic opinion pooling view to multimodal VAEs, we aim to design a pooling function that aggregates unimodal posteriors to approximate the true joint posterior. MVAE~\citep{wu2018multimodal} employs a PoE, corresponding to log-linear pooling, which yields sharp approximations but struggles to optimize the individual experts, as noted by the authors. In contrast, MMVAE~\citep{shi2019variational} adopts a MoE, corresponding to linear pooling, which optimizes experts effectively but cannot produce a sharper posterior than its components. Thus, the choice of pooling function directly shapes the properties of the learned model. Moreover, as highlighted by~\citet{winkler1981combining, dietrich2016probabilistic, mancisidor2025aggregation}, both PoE and MoE depend on an overoptimistic independence assumption between experts for simplicity. In practice, this assumption rarely holds, as data modalities are different sources of information about the same object. By contrast, H\"older pooling with $\alpha=0.5$ introduces cross terms such as $\sqrt{q_i(\vz)q_j(\vz)}$, which naturally induce soft dependencies between experts by amplifying regions of mutual support.

\begin{figure}[t!]
\begin{center}
\includegraphics[width=0.48\textwidth]{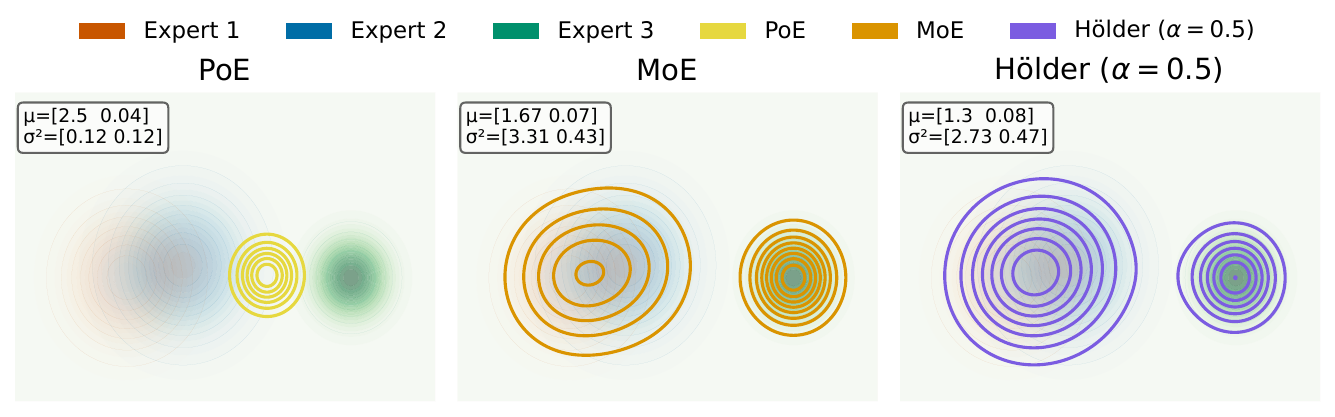}
\end{center}
\vspace{-0.1in}
\caption{ PoE, MoE, and H\"older pooling ($\alpha=0.5$) with two agreeing experts and one disagreeing sharp expert.}
\label{fig:dist_experts}
\vspace{-0.1in}
\end{figure}

To illustrate, we consider three Gaussian experts with diagonal covariance  $\{\mathcal{N}(\vmu_j, \mathrm{diag}(\vsigma_j^2))\}_{j=1}^3$, expert~1 with $\vmu_1 = (0, 0)^\top, \vsigma_1^2 = (0.5, 0.5)^\top$, and expert~2 with $\vmu_2 = (1.0, 0.2)^\top, \vsigma_2^2 = (0.6, 0.6)^\top$, close to each other, while expert~3 with $\vmu_3 = (4, 0)^\top, \vsigma_3^2 = (0.2, 0.2)^\top$, is sharp and far away. As shown in Figure~\ref{fig:dist_experts}, PoE shifts toward the sharpest expert, while MoE spreads mass across all experts and fails to capture the local agreement between experts~1~and~2. In contrast, H\"older pooling with $\alpha=0.5$ overcomes this by placing more mass around the first two experts, yielding a sharper distribution with reduced variance compared to MoE. This shows how H\"older pooling better reflects multimodal structure than either PoE or MoE.


In the multimodal VAE setting, each unimodal posterior (expert) is modeled as a multivariate Gaussian with diagonal covariance matrix, which we propose to aggregate using the Hellinger function. This function is derived from H\"older pooling with $\alpha = 0.5$ and uses moment matching~\citep{bowman2004estimation} to project the pooled density onto a diagonal Gaussian. In the next part, we derive this aggregation function and then use it in the proposed HELVAE.

\paragraph{Hellinger aggregation.} Let unimodal inference distributions $\{q_{\phi_\subindex}\}_{\subindex=1}^\modalities$ be explicitly given as Gaussian probability densities on $\sR^D$ with diagonal covariance matrices, $\{ \mathcal{N}(\vmu_\subindex, \mathrm{diag}(\vsigma_\subindex^2)) \}_{\subindex=1}^\modalities$, where $\vmu_\subindex = (\mu_{\subindex,1}, \mu_{\subindex,2}, \dots, \mu_{\subindex,D})^\top \in \sR^D$ and
$\vsigma_\subindex^2 = (\sigma_{\subindex,1}^2, \sigma_{\subindex,2}^2, \dots, \sigma_{\subindex,D}^2)^\top \in \sR^D$. 
For each pair of modalities $(i,j)$ with $1 \leq i < j \leq \modalities$, we then define the auxiliary parameters $\vmu_{ij}, \vsigma_{ij}^2 \in \sR^D$ by specifying their corresponding dimension $d \in \{1, 2, \dots, D\}$ as follows:
\begin{equation*}
    \mu_{ij,d} = \frac{\mu_{i,d}\,\sigma_{j,d}^2 + \mu_{j,d}\,\sigma_{i,d}^2}{\sigma_{i,d}^2 + \sigma_{j,d}^2}, 
    \quad
    \sigma_{ij,d}^2 = \frac{2\sigma_{i,d}^2 \sigma_{j,d}^2}{\sigma_{i,d}^2 + \sigma_{j,d}^2}.
\end{equation*}
The Bhattacharyya coefficient~\footnote{The Bhattacharyya coefficient measures the overlap between two probability distributions, taking values in $[0,1]$, with $1$ indicating identical distributions.} between $q_{\phi_i}$ and $q_{\phi_j}$ is given by
\begin{equation*}
    S_{ij} = \prod_{d=1}^D \sqrt{\frac{2\sigma_{i,d}\sigma_{j,d}}{\sigma_{i,d}^2 + \sigma_{j,d}^2}} 
    \exp\!\left(-\frac{(\mu_{i,d} - \mu_{j,d})^2}{4(\sigma_{i,d}^2 + \sigma_{j,d}^2)}\right),
\end{equation*}
which allows us to compute the  the normalizing constant as:
\begin{equation*}
    c = \modalities + 2 \sum_{i=1}^M \sum_{j>i}^M S_{ij}.
\end{equation*}

Finally, we approximate the joint posterior resulting from H\"older pooling with $\alpha=0.5$ in Equation~\ref{eqn:holder} via moment matching as a $D$-dimensional Gaussian distribution, $q_\phi(\vz|\sX) \sim \mathcal{N}(\tilde{\vmu}, \tilde{\mSigma})$, with mean $\tilde{\vmu} \in \sR^D$ and covariance matrix $\tilde{\mSigma} \in \sR^{D \times D}$, corresponding to the first moment and the second central moment of the H\"older-pooled density, i.e., 
    \begin{equation*}
        \tilde{\vmu} = \int \vz q(\vz)d\vz, \quad \tilde{\mSigma} = \int \vz \vz^\top q(\vz)d\vz - \tilde{\vmu} \tilde{\vmu}^\top.
    \end{equation*}
    In VAE models, the latent variable is typically parameterized as a multivariate Gaussian with diagonal covariance~\citep{kingma2013auto}. Thus, we simplify $\tilde{\mSigma}$ by removing off-diagonal terms, yielding $\tilde{\mSigma} \approx \mathrm{diag}(\tilde{\sigma}_1^2, \tilde{\sigma}_2^2, \dots, \tilde{\sigma}_D^2)$. 
As a result, we  compute  the mean and the variance of each dimension $d \in \{1,2,\dots,D\}$ of the latent vector $\mathbf{z}$  as follows:
\begin{equation*}
    \tilde{\mu}_d = \frac{1}{c}\left(\sum_{j=1}^M \mu_{j, d} + 2 \sum_{i=1}^M \sum_{j>i}^M \mu_{ij, d} S_{ij}\right),
\end{equation*}
\begin{equation*}
\begin{split}
\tilde{\sigma}_d^2
&= \frac{1}{c}\Bigg(
      \sum_{j=1}^M (\mu_{j,d}^2 + \sigma_{j,d}^2) \\
&\quad {}+{} 2 \sum_{i=1}^M \sum_{j>i}^M
      (\mu_{ij,d}^2 + \sigma_{ij,d}^2) S_{ij}
    \Bigg)
 - (\tilde{\mu}_d)^2.
\end{split}
\end{equation*}

    

The derivation above presents the moment-matching approximation of H\"older pooling with $\alpha=0.5$ for a set of multivariate Gaussians with diagonal covariance. The complete derivation is given in Appendix~\ref{proof:lem_hel}, where it is derived under the assumption of subset weights. Since determining these weights in multimodal VAEs is generally nontrivial, we follow PoE and MoE and set them \textit{uniformly}. \emph{Hellinger aggregation defines a novel multimodal VAE, called HELVAE.} An important property of HELVAE is that it does not rely on sub-sampling, which negatively affects likelihood estimation and generative quality in multimodal VAEs. As  shown by our experiments in Section~\ref{sec:experiments}, HELVAE leads to better trade-offs between generation quality and coherence than MVAE and MMVAE, yielding competitive or even better results than more complex approaches such as the state-of-the-art, MMVAE+. 

\begin{figure}[t!]
\begin{center}
\includegraphics[width=0.48\textwidth]{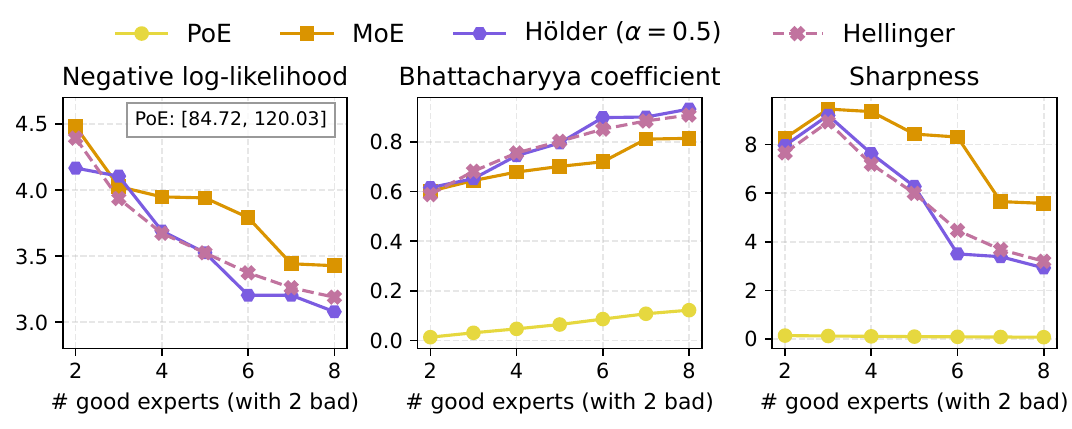}
\end{center}
\vspace{-0.1in}
\caption{ Performance of PoE, MoE, H\"older pooling ($\alpha=0.5$), and Hellinger aggregators as a function of the number of good experts (with two bad experts fixed). Evaluation is based on negative log-likelihood ($\downarrow$), the Bhattacharyya coefficient ($\uparrow$), and sharpness ($\downarrow$). We show the minimum and maximum NLL values for PoE.}
\label{fig:good_experts}
\vspace{-0.1in}
\end{figure}

\paragraph{Illustrative example.} We construct a synthetic setup to evaluate PoE, MoE, H\"older ($\alpha=0.5$), and Hellinger in Figure~\ref{fig:good_experts}. The set of experts is defined as:
\begin{equation*}
q_i(\vz) = 
\begin{cases}
\mathcal{N}(\vz \mid (0,0)^\top, 0.5\mI), & i \in \mathcal{G} \text{ (good experts)}, \\
\mathcal{N}(\vz \mid (4, 4)^\top, 0.2\mI), & i \in \mathcal{B} \text{ (bad experts)},
\end{cases}
\end{equation*}
and the true distribution is $\mathcal{N}(\mathbf{0}, \mI)$. Evaluation metrics include (i) expected negative log-likelihood (NLL), estimated by Monte Carlo samples from the true distribution; (ii) the Bhattacharyya coefficient with respect to the true distribution; and (iii) sharpness, measured as the trace of the covariance of the Gaussian approximation. Since MoE and H\"older ($\alpha=0.5$) do not yield Gaussian densities in closed form, all metrics are approximated by sampling. The results show that PoE performs poorly, with high NLL and low Bhattacharyya coefficient, as it is biased toward the sharper bad expert. Other models maintain low NLL. Compared to MoE, H\"older ($\alpha=0.5$) and Hellinger yield lower NLL, sharper distributions and stronger distributional overlap with the true distribution as the number of good experts increases, demonstrating their ability to capture dependencies among good experts. 

\paragraph{Mixture of HELVAEs.} Following MoPoE~\citep{sutter2021generalized}, we introduce a variant of HELVAE that constructs a mixture of H\"older poolings with $\alpha = 0.5$, termed MoHELVAE. Specifically, we consider the powerset of $M$ modalities $\mathcal{P}(\sX)$, consisting of all $2^M - 1$ non-empty subsets. For each subset $\sX_k \in \mathcal{P}(\sX)$, we define its posterior approximation as:
    \begin{equation*}
        \textstyle \tilde{q}_\phi(\vz|\sX_k) = \text{HEL} (\{q_{\phi_j}(\vz|\vx_j), \forall \vx_j \in \sX_k\}),
    \end{equation*}
    then the joint posterior of MoHELVAE is given by 
    \begin{equation*}
        q_\phi(\vz|\sX) = \frac{1}{2^M - 1} \sum_{\sX_k \in \mathcal{P}(\sX)} \tilde{q}_\phi(\vz|\sX_k).
    \end{equation*}
The proposed MoHELVAE model outperforms state-of-the-art approaches in terms of latent representation and generative coherence on the challenging CelebA dataset, as demonstrated in Ablation study~\ref{ab:mohel}.


\section{EXPERIMENTAL RESULTS}
\label{sec:experiments}

\begin{figure*}[t!]
\begin{center}
\includegraphics[width=\textwidth]{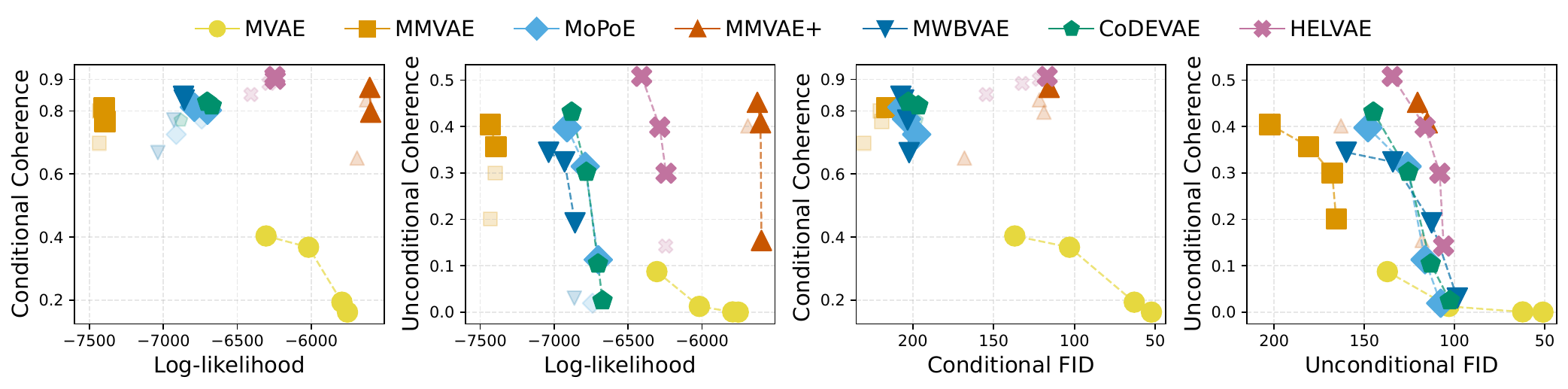}
\end{center}
\vspace{-0.15in}
\caption{ Trade-offs on the PolyMNIST dataset between generative coherence ($\uparrow$) and log-likelihood estimation ($\uparrow$), as well as between generative coherence and generative quality ($\downarrow$), for $\beta \in \{1, 2.5, 5, 10\}$. For each model, the Pareto front (dashed line) connects the non-dominated points that achieve the best trade-offs.}
\label{fig:pm_coh_fid}
\vspace{-0.05in}
\end{figure*}

\begin{figure*}[t!]
\begin{center}
\includegraphics[width=0.97\textwidth]{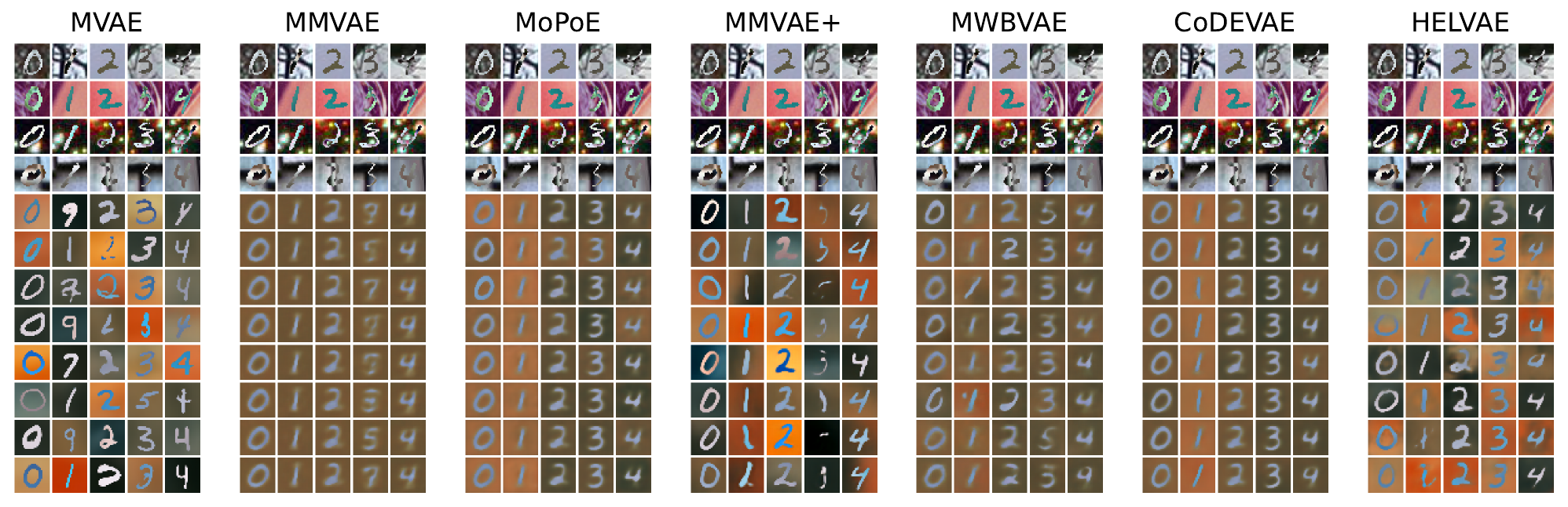}
\end{center}
\vspace{-0.1in}
\caption{Conditionally generated samples of the second modality (fifth to last rows), given the corresponding test examples from the other four modalities (first four rows) on PolyMNIST, with each model evaluated at its best $\beta \in \{1, 2.5, 5, 10\}$.}
\label{fig:pm_4_1}
\vspace{-0.1in}
\end{figure*}

Following prior work in this domain~\citep{sutter2021generalized, daunhawer2021limitations, palumbo2023mmvae+}, we evaluate our approach on three benchmark datasets: PolyMNIST~\citep{sutter2021generalized}, Caltech Birds (CUB) Image-Captions~\citep{wah2011caltech}, and bimodal CelebA~\citep{liu2015deep}. PolyMNIST is a synthetic dataset with five modalities, where each sample consists of MNIST digits of the same label patched onto random crops from five distinct background images. CUB Image-Captions consists of bird images paired with textual descriptions. The bimodal CelebA dataset consists of human face images paired with text describing facial attributes.

We compare the performance of HELVAE with MVAE~\citep{wu2018multimodal}, MMVAE~\citep{shi2019variational}, MoPoE~\citep{sutter2021generalized}, MMVAE+~\citep{palumbo2023mmvae+}, MWBVAE (Mixture of WBVAE)~\citep{qiu2025multimodal}, and CoDEVAE~\citep{mancisidor2025aggregation}. Among these, MMVAE+ is the only model that incorporates both modality-specific and shared latent variables, while the others focus on aggregating unimodal posterior distributions.

We evaluate performance based on generative coherence and generative quality (measured using FID~\citep{heusel2017gans}), using samples generated from both the joint posterior and the prior. To assess the approximation quality of the joint posterior, we estimate the log-likelihood, which provides a tightness lower bound. We further evaluate the quality of the joint latent representation $\vz$ using a linear classifier. All experiments report the average performance over three different seeds. Technical details and further results are provided in Appendix~\ref{app:exp}.

\begin{table*}[t!]
\caption{ Generative coherence and quality on the bimodal CelebA dataset, with each model evaluated at its best $\beta \in \{1, 2.5, 5\}$, and rankings are reported accordingly.}
\vspace{-0.1in}
\label{tab:celeba_coh_fid}
\begin{center}
\begin{tabular}{rcccc}
\toprule
& \multicolumn{2}{c}{Conditional} & \multicolumn{2}{c}{Unconditional} \\
& Coherence ($\uparrow$) & FID ($\downarrow$) & Coherence ($\uparrow$) & FID ($\downarrow$) \\
\midrule
MVAE          & 0.315 {\tiny $\pm$ 0.013 (7)} & \textbf{75.62 {\tiny $\pm$ 3.01 (1)}}  & 0.199 {\tiny $\pm$ 0.021 (6)} & \textbf{78.64 {\tiny $\pm$ 2.73 (1)}} \\
MMVAE           & 0.370 {\tiny $\pm$ 0.007 (6)} & 95.10 {\tiny $\pm$ 11.68 (6)} & 0.209 {\tiny $\pm$ 0.011 (4)} & 91.04 {\tiny $\pm$ 5.82 (6)} \\
MoPoE    & 0.376 {\tiny $\pm$ 0.005 (4)} & 81.74 {\tiny $\pm$ 2.85 (3)}  & 0.219 {\tiny $\pm$ 0.006 (2)} & 82.50 {\tiny $\pm$ 2.45 (3)} \\
MMVAE+ & 0.382 {\tiny $\pm$ 0.013 (2)} & 99.72 {\tiny $\pm$ 5.23 (7)}  & \textbf{0.258 {\tiny $\pm$ 0.024 (1)}} & 101.85 {\tiny $\pm$ 5.13 (7)} \\
MWBVAE   & 0.376 {\tiny $\pm$ 0.003 (4)} & 88.28 {\tiny $\pm$ 1.58 (4)}  & 0.197 {\tiny $\pm$ 0.006 (7)} & 88.74 {\tiny $\pm$ 1.09 (5)} \\
CoDEVAE & 0.378 {\tiny $\pm$ 0.004 (3)} & 91.24 {\tiny $\pm$ 11.27 (5)} & 0.219 {\tiny $\pm$ 0.020 (2)} & 88.15 {\tiny $\pm$ 6.68 (4)} \\
\textbf{HELVAE}     & \textbf{0.386 {\tiny $\pm$ 0.003 (1)}} & 80.17 {\tiny $\pm$ 1.52 (2)} & 0.204 {\tiny $\pm$ 0.013 (5)} & 81.89 {\tiny $\pm$ 1.56 (2)} \\
\bottomrule
\end{tabular}
\end{center}
\vspace{-0.15in}
\end{table*}

\subsection{Trade-offs between generative coherence and generative quality}

\paragraph{PolyMNIST.} Figure~\ref{fig:pm_coh_fid} shows the trade-offs between generative coherence and log-likelihood estimation, as well as between generative coherence and quality across all $\beta$ values~\footnote{$\beta$-VAE~\citep{higgins2017beta} scales the $\KL$ term to balance reconstruction and disentanglement.}. MVAE, based on PoE, yields tight log-likelihood estimates but performs poorly in coherence. In contrast, mixture-based models such as MMVAE, MoPoE, MWBVAE, and CoDEVAE achieve relatively high coherence, though not the best overall. Meanwhile, MMVAE+ and HELVAE both achieve better trade-offs:  HELVAE attains higher coherence in both conditional and unconditional settings; while MMVAE+ provides a tighter log-likelihood bound, which is a direct result of the modality-specific variables. While generative coherence is important, the generated modalities should also be of high quality. Following the same pattern, MVAE generates high-quality images but lacks coherence. HELVAE achieves generative quality comparable to MMVAE+ while attaining better coherence. The other models yield lower-quality images due to sub-sampling of modalities during training. Qualitative results in Figure~\ref{fig:pm_4_1} show that HELVAE achieves a better balance between coherence and quality compared to the other methods.

\begin{table}[t!]
\caption{Caption-to-image conditional generative coherence and quality on the CUB Image-Captions dataset, with each model evaluated at its best $\beta \in \{1, 2.5, 5\}$, and rankings for each metric are reported accordingly.}
\vspace{-0.1in}
\label{tab:cub_coh_fid}
\begin{center}
\begin{tabular}{rcc}
\toprule
& \multicolumn{2}{c}{Conditional} \\
& Coherence ($\uparrow$) & FID ($\downarrow$) \\
\midrule
MVAE    & 0.233 {\tiny $\pm$ 0.038 (7)}     & \textbf{148.69 {\tiny $\pm$ 1.97 (1)}} \\
MMVAE   & 0.688 {\tiny $\pm$ 0.054 (5)}     & 225.77 {\tiny $\pm$ 1.66 (7)} \\
MoPoE   & 0.675 {\tiny $\pm$ 0.084 (6)}     & 202.19 {\tiny $\pm$ 4.68 (6)} \\
MMVAE+  & 0.720 {\tiny $\pm$ 0.090 (3)}     & 164.94 {\tiny $\pm$ 1.50 (3)} \\
MWBVAE  & 0.704 {\tiny $\pm$ 0.079 (4)}     & 196.42 {\tiny $\pm$ 5.58 (5)} \\
CoDEVAE & \textbf{0.750 {\tiny $\pm$ 0.050 (1)}}     & 175.97 {\tiny $\pm$ 0.30 (4)} \\
\textbf{HELVAE} & \textbf{0.750 {\tiny $\pm$ 0.105 (1)}}     & 157.56 {\tiny $\pm$ 0.95 (2)} \\
\bottomrule
\end{tabular}
\end{center}
\vspace{-0.05in}
\end{table}

\begin{table}[t!]
\caption{ Classification accuracy based on the latent representation learned from the bimodal CelebA dataset, with each model evaluated at its best $\beta \in \{1, 2.5, 5\}$. We report the mean average precision over all attributes (I: Image; T: Text; Joint: I and T), and rankings accordingly.}
\vspace{-0.1in}
\label{tab:lr_coh_results}
\begin{center}
\begin{tabular}{rccc}
\toprule
& \multicolumn{3}{c}{Latent Representation ($\uparrow$)} \\
& I & T & Joint \\
\midrule
MVAE      & 0.326 {\tiny (6)} & 0.327 {\tiny (7)} & 0.349 {\tiny (6)} \\
MMVAE     & 0.360 {\tiny (5)} & 0.410 {\tiny  (4)} & 0.369 {\tiny (5)} \\
MoPoE     & 0.362 {\tiny (3)} & 0.398 {\tiny (6)} & 0.398 {\tiny (4)} \\
MMVAE+ & 0.314 {\tiny (7)} & \textbf{0.461 {\tiny (1)}} & 0.346 {\tiny (7)} \\
MWBVAE    & 0.361 {\tiny (4)} & 0.415 {\tiny (3)} & \textbf{0.410 {\tiny (1)}} \\
CoDEVAE   & \textbf{0.365 {\tiny(1)}} & 0.418 {\tiny (2)} & 0.399 {\tiny (3)} \\
\textbf{HELVAE}   & \textbf{0.365 {\tiny (1)}} & 0.400 {\tiny (5)} & 0.408 {\tiny (2)}\\
\bottomrule
\end{tabular}
\end{center}
\vspace{-0.15in}
\end{table}

\paragraph{CUB Image-Captions.} Table~\ref{tab:cub_coh_fid} reports quantitative results for caption-to-image generation, where conditional coherence is computed following~\citet{palumbo2023mmvae+}, since we do not have annotations for the shared content between modalities. Results from MVAE, MMVAE, and MoPoE highlight the difficulty of the task with real images: MVAE produces reasonable image quality but lacks coherence, while MMVAE and MoPoE suffer from modality collapse in mixture-based models. In contrast, our model achieves the highest conditional coherence, which is the same as CoDEVAE but with lower conditional FID than other models. HELVAE ranks second in FID and outperforms MMVAE+. These results confirm  the effectiveness of our approach in real-world multimodal settings with substantial modality-specific information and significant heterogeneity across modalities.

\begin{figure*}[t!]
    \centering
    \begin{subfigure}{0.13\textwidth}
        \includegraphics[width=\linewidth]{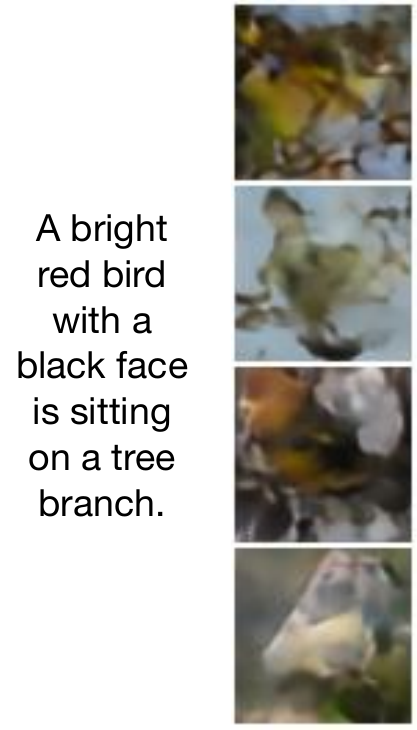}
        \caption{ MVAE}
    \end{subfigure}
    \hfill
    \begin{subfigure}{0.13\textwidth}
        \includegraphics[width=\linewidth]{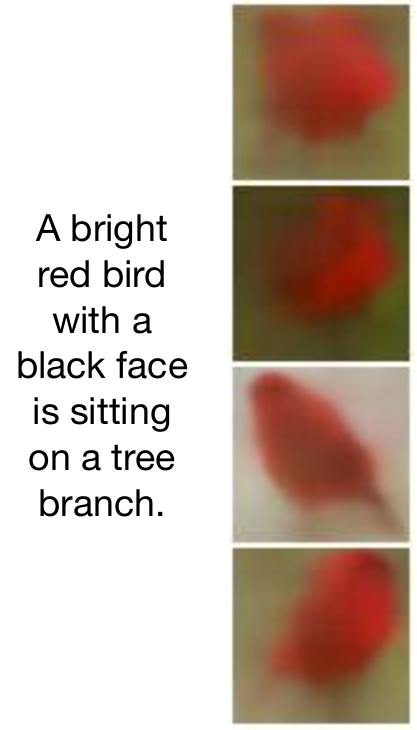}
        \caption{ MMVAE}
    \end{subfigure}
    \hfill
    \begin{subfigure}{0.13\textwidth}
        \includegraphics[width=\linewidth]{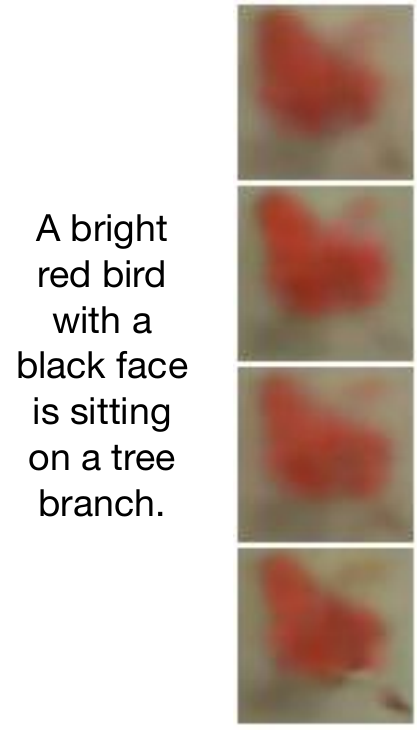}
        \caption{ MoPoE}
    \end{subfigure}
    \hfill
    \begin{subfigure}{0.13\textwidth}
        \includegraphics[width=\linewidth]{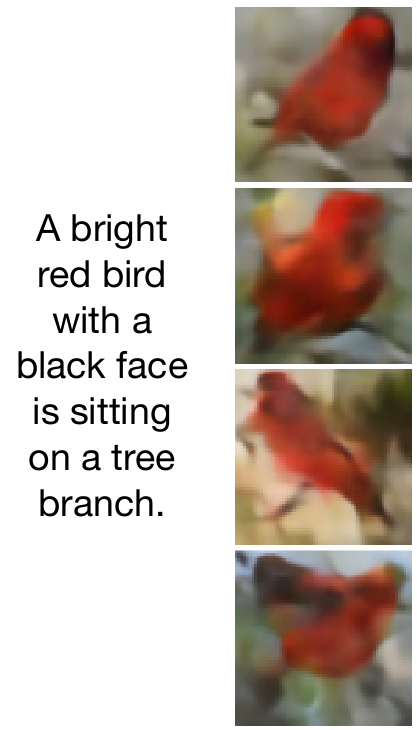}
        \caption{ MMVAE+}
    \end{subfigure}
    \hfill
    \begin{subfigure}{0.13\textwidth}
        \includegraphics[width=\linewidth]{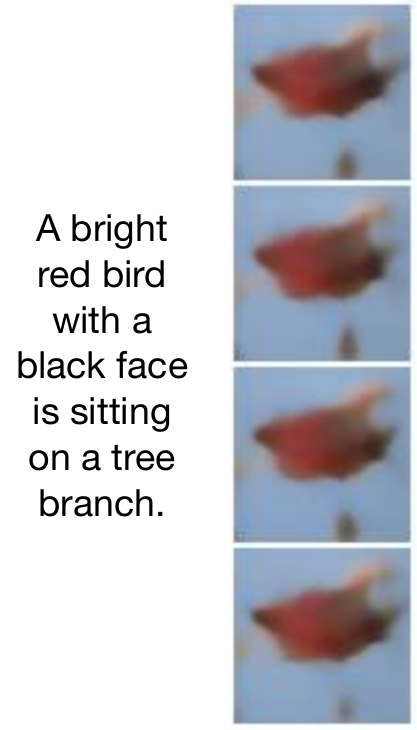}
        \caption{ MWBVAE}
    \end{subfigure}
    \hfill
    \begin{subfigure}{0.13\textwidth}
        \includegraphics[width=\linewidth]{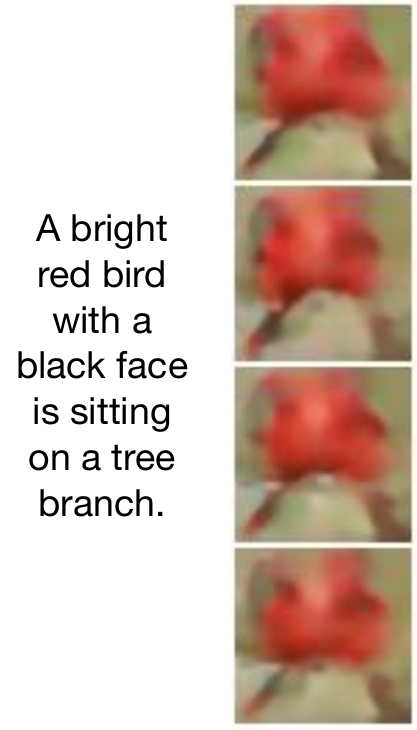}
        \caption{ CoDEVAE}
    \end{subfigure}
    \hfill
    \begin{subfigure}{0.13\textwidth}
        \includegraphics[width=\linewidth]{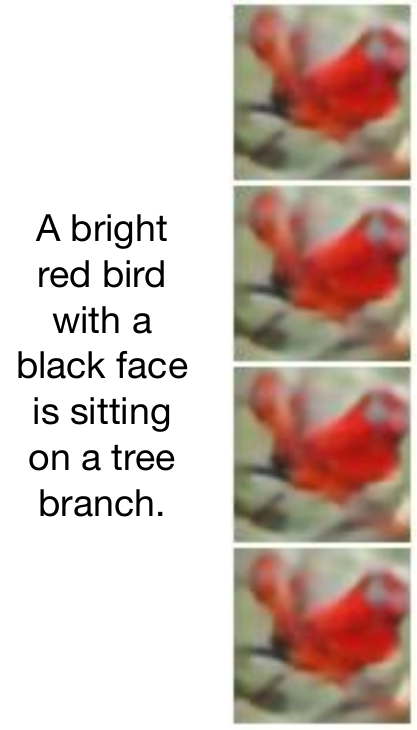}
        \caption{ HELVAE}
    \end{subfigure}
    \vspace{-0.05in}
    \caption{ Qualitative results for caption-to-image generation on the CUB Image-Captions dataset, with each model evaluated at its best $\beta \in \{1, 2.5, 5\}$, where the input caption is used to conditionally generate four images per model.}
    \vspace{-0.05in}
    \label{fig:cub_text_to_img_1}
\end{figure*}

Figure~\ref{fig:cub_text_to_img_1} shows qualitative results for caption-to-image generation. As we can see, although MVAE achieves high-quality images, these images do not follow the captions well and are noisy, resulting in low coherence. MMVAE, MWBVAE, and CoDEVAE often collapse modality-specific features to average values, yielding blurred images. In contrast, HELVAE and MMVAE+ avoid this collapse, producing images that are both high quality and coherent with the captions. Although HELVAE images are less diverse due to the lack of a modality-specific latent space, its generated images are well-shaped and align closely with the captions.

\begin{figure*}[t!]
\begin{center}
\includegraphics[width=\textwidth]{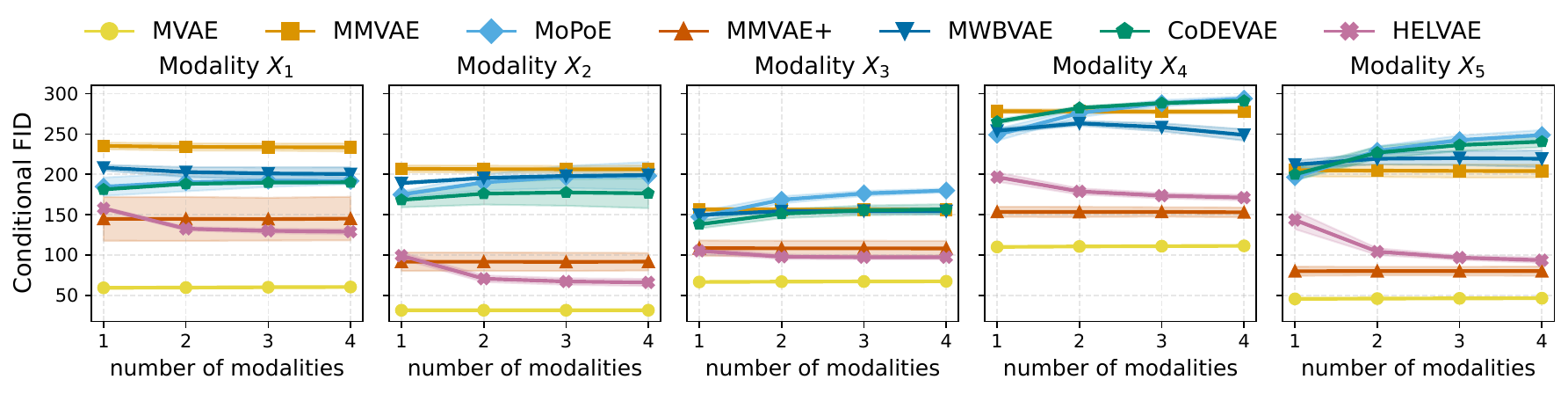}
\end{center}
\vspace{-0.15in}
\caption{ Conditional generative quality ($\downarrow$) on the PolyMNIST dataset for target modalities $X_1 \dots X_5$ as a function of the number of input modalities, averaged across subsets of a given size excluding the target. Markers indicate means, and shaded regions represent standard deviations.}
\label{fig:pm_fid_num_mods}
\vspace{-0.1in}
\end{figure*}

\paragraph{Bimodal CelebA.} As shown in Table~\ref{tab:celeba_coh_fid}, HELVAE achieves the highest conditional coherence while maintaining generative quality (second-best FID). By contrast, MMVAE+ attains the highest unconditional coherence but suffers from high FID. CoDEVAE and MoPoE achieve the second-best unconditional coherence; however, CoDEVAE has high FID, and MoPoE underperforms HELVAE on other metrics. Consistent with the previous dataset, MVAE has the best generative quality but at high degradation in coherence.

\subsection{Latent representation quality}
The quality of the learned representations serves as a proxy for their usefulness in downstream tasks. From Table~\ref{tab:lr_coh_results}, when evaluating latent representation accuracy on the CelebA dataset, HELVAE shares the highest score on the image modality with CoDEVAE. Although it ranks fifth on the text modality, it achieves the second-best performance in the joint setting, showing a gain over its single-modality performance. In contrast, MWBVAE, CoDEVAE, and MMVAE+ do not benefit from additional modalities in the same way. This is an important criterion in multimodal generative models, known as synergy~\citep{shi2019variational}.



\section{ABLATION STUDIES}

\begin{table}[t!]
\caption{ Model performance of HELVAE and MoHELVAE on the bimodal CelebA dataset. Evaluation is based on generative coherence (Coh $\uparrow$), generative quality (FID $\downarrow$), and latent representation accuracy (LR $\uparrow$).}
\vspace{-0.1in}
\label{tab:mohel}
\begin{center}
\begin{tabular}{llcc}
\toprule
& & HELVAE & MoHELVAE \\
\midrule
\multirow{2}{*}{\rot{360}{Coh}} 
  & Conditional   & 0.386 {\tiny $\pm$ 0.003} & \textbf{0.394 {\tiny $\pm$ 0.011}} \\
  & Unconditional & 0.204 {\tiny $\pm$ 0.013} & \textbf{0.229 {\tiny $\pm$ 0.020}} \\
\midrule
\multirow{2}{*}{\rot{360}{FID}} 
  & Conditional   & \textbf{80.17 {\tiny $\pm$ 1.52}} & 82.28 {\tiny $\pm$ 4.72}  \\
  & Unconditional & \textbf{81.89 {\tiny $\pm$ 1.56}} & 83.14 {\tiny $\pm$ 4.87}  \\
\midrule
\multirow{3}{*}{\rot{360}{LR}} 
  & Image          & 0.365 {\tiny $\pm$ 0.004} & \textbf{0.379 {\tiny $\pm$ 0.013}} \\
  & Text           & 0.400 {\tiny $\pm$ 0.004} & \textbf{0.441 {\tiny $\pm$ 0.019}} \\
  & Image \& Text     & 0.408 {\tiny $\pm$ 0.005} & \textbf{0.430 {\tiny $\pm$ 0.018}} \\
\bottomrule
\end{tabular}
\end{center}
\vspace{-0.15in}
\end{table}

\subsection{Effectiveness of Mixture of HELVAEs}
\label{ab:mohel}
From Table~\ref{tab:mohel}, MoHELVAE surpasses HELVAE in generative coherence (conditional and unconditional) and latent representation accuracy. However, its FID is slightly higher than that of HELVAE, and representation accuracy drops when combining image and text, a pattern also observed in other mixture-based models on the bimodal CelebA dataset. Compared with Tables~\ref{tab:celeba_coh_fid} and~\ref{tab:lr_coh_results}, MoHELVAE outperforms other models and, in particular, achieves a better balance between coherence and quality than MMVAE+, highlighting the efficiency of incorporating the joint posterior across all subsets of modalities.

\subsection{Effect of the number of input modalities on model performance}
Figure~\ref{fig:pm_fid_num_mods} shows PolyMNIST performance as a function of the number of observed input modalities at $\beta = 2.5$, the setting recommended for MMVAE+~\citep{palumbo2023mmvae+}. As the number of observed modalities increases, MoPoE and CoDEVAE exhibit decreased generative quality, while MMVAE+ maintains competitive FID scores but shows only limited benefit from additional inputs. In contrast, HELVAE exhibits clear multimodal synergy: its FID steadily decreases as additional modalities are provided, suggesting that it can effectively exploit complementary information across views. This trend is also consistent with the qualitative results: for examples showing that HELVAE generates more coherent samples when conditioned on four modalities rather than one, see Appendix~\ref{app:exp}.

\begin{figure*}[t!]
\begin{center}
\includegraphics[width=\textwidth]{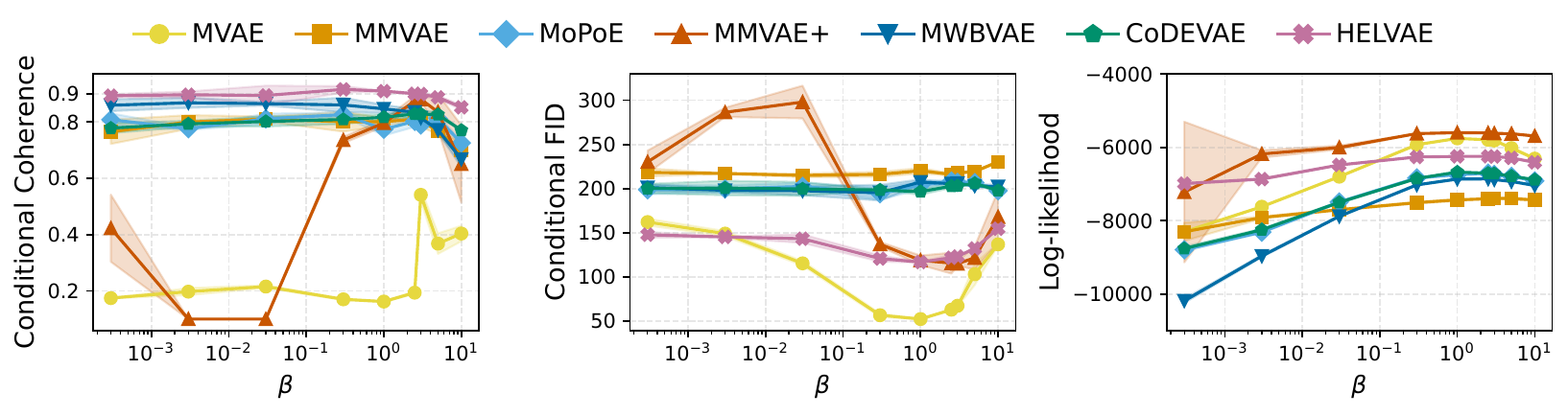}
\end{center}
\vspace{-0.15in}
\caption{Performance on the PolyMNIST dataset for different values of $\beta \in \{3e^{-4}, 3e^{-3}, 3e^{-2}, 3e^{-1}, 1, 2.5, 3, 5, 10\}.$
Metrics include conditional generative coherence ($\uparrow$), conditional generative quality ($\downarrow$), and log-likelihood estimation ($\uparrow$). Markers denote means, while shaded regions represent standard deviations.}
\label{fig:pm_beta}
\vspace{-0.1in}
\end{figure*}

\subsection{Effect of $\beta$ on model performance}
Figure~\ref{fig:pm_beta} reports performance across different $\beta$ values on the PolyMNIST dataset, chosen following~\citet{daunhawer2021limitations}. While MMVAE+ reaches the highest log-likelihood, its coherence drops and FID increases sharply at both small and large $\beta$. MMVAE, MoPoE, MWBVAE, and CoDEVAE remain stable but yield lower coherence and higher FID than HELVAE. Overall, HELVAE delivers the best performance. It achieves the highest coherence, maintains competitive FID, and attains high log-likelihood across a broad range of $\beta$ values, indicating that its performance is robust to the choice of this hyperparameter.

\subsection{Computational cost analysis}

\begin{table}[t!]
\caption{Average training time per batch, measured in seconds, on the three benchmark datasets.}
\vspace{-0.1in}
\label{tab:training_time}
\begin{center}
\begin{tabular}{rccc}
\toprule
 & \makecell[c]{PolyMNIST\\$M=5$} & \makecell[c]{CUB\\$M=2$} & \makecell[c]{CelebA\\$M=2$} \\
\midrule
MVAE    & 0.1887 & 0.2184 & 1.0294 \\
MMVAE   & 0.2362 & 0.1876 & 0.5764 \\
MoPoE   & 0.1216 & 0.2269 & 0.6011 \\
MMVAE+  & 0.2884 & 0.2270 & 0.9916 \\
MWBVAE  & 0.1134 & 0.1408 & 0.5712 \\
CoDEVAE & 0.1614 & 0.0950 & 0.5817 \\
\textbf{HELVAE}  & \textbf{0.0895} & \textbf{0.0863} & \textbf{0.5272} \\
\textbf{MoHELVAE} & ---    & ---    & 0.5918 \\
\bottomrule
\end{tabular}
\end{center}
\vspace{-0.1in}
\end{table}

Table~\ref{tab:training_time} reports the average batch training time on the three benchmark datasets. HELVAE consistently has the shortest batch training time across all three. For mixture-based models with $\mathcal{O}(2^M)$ cost for the aggregation, such as MoPoE, MWBVAE, CoDEVAE, and MoHELVAE, training times on CelebA are comparable to HELVAE, but the gaps are larger on PolyMNIST and CUB. By contrast, MVAE, MMVAE, and MMVAE+ are the most time-consuming models: for MVAE, this is mainly due to its training scheme rather than PoE aggregation, while for MMVAE and MMVAE+, the higher cost comes from the multi-sample estimator. Overall, HELVAE is the most computationally efficient baseline while remaining effective.

Besides, \textit{when the number of modalities is small}, such as in CelebA with two modalities, MoHELVAE remains computationally efficient, with batch training time comparable to HELVAE and MoPoE, while also outperforming other approaches in terms of latent representation quality and generative coherence.


\section{CONCLUSION}


This work introduced HELVAE, a multimodal VAE based on Hellinger aggregation and H\"older pooling, that improves semantic coherence without sacrificing generative quality. However, the gains in conditional coherence come with a slight trade-off in unconditional coherence and sample diversity, especially compared to MMVAE+, which uses modality-specific private latents. A natural promising extension is therefore to incorporate private latent variables into HELVAE. Moreover, we plan to study HELVAE with more expressive encoders, such as CLIP-style encoders~\citep{radford2021learning}. In addition, learnable pooling weights or the use of negative weights, as proposed in the squared subtractive mixture model~\citep{loconte2023subtractive, loconte2025sum, loconte2025square}, may further improve performance. We also aim to extend Hellinger aggregation beyond VAEs, including to Gaussian process experts~\citep{pmlr-v119-cohen20b} and other broader settings that aggregate information from expert distributions~\citep{liusie2024efficient, zhang2025product}. In the future, we want to explore more multimodal learning frameworks with
other powerful generative models, such as diffusion~\citep{chen2024diffusion, rojas2025diffuse} and flows~\citep{campbell2024generative}. The Pytorch code of our paper can be found at \href{https://github.com/vothuckhanhhuyen/hellinger-multimodal-variational-autoencoders}{https://github.com/vothuckhanhhuyen/hellinger-multimodal-variational-autoencoders}.


\subsubsection*{Acknowledgements}

We thank the anonymous reviewers at AISTATS 2026 for their helpful and constructive comments. We also thank María Martínez García and Batuhan Koyuncu from the Probabilistic ML group at Saarland University for valuable discussions. This work has been supported by the project \textit{“Society-Aware Machine Learning: The paradigm shift demanded by society to trust machine learning”}, funded by the European Union and led by Isabel Valera (ERC-2021-STG, SAML, 101040177). Huyen Vo also acknowledges her membership in the CS@Max Planck doctoral program. Views and opinions expressed are, however, those of the author(s) only and do not necessarily reflect those of the aforementioned funding agencies. Neither of the aforementioned parties can be held responsible for them.


\bibliography{aistats2026}





\section*{Checklist}



\begin{enumerate}

  \item For all models and algorithms presented, check if you include:
  \begin{enumerate}
    \item A clear description of the mathematical setting, assumptions, algorithm, and/or model. Yes
    \item An analysis of the properties and complexity (time, space, sample size) of any algorithm. Yes
    \item (Optional) Anonymized source code, with specification of all dependencies, including external libraries. Yes. See Appendix
  \end{enumerate}

  \item For any theoretical claim, check if you include:
  \begin{enumerate}
    \item Statements of the full set of assumptions of all theoretical results. Yes
    \item Complete proofs of all theoretical results. Yes. See Appendix
    \item Clear explanations of any assumptions. Yes     
  \end{enumerate}

  \item For all figures and tables that present empirical results, check if you include:
  \begin{enumerate}
    \item The code, data, and instructions needed to reproduce the main experimental results (either in the supplemental material or as a URL). Yes. See Appendix
    \item All the training details (e.g., data splits, hyperparameters, how they were chosen). Yes. See Appendix
    \item A clear definition of the specific measure or statistics and error bars (e.g., with respect to the random seed after running experiments multiple times). Yes
    \item A description of the computing infrastructure used. (e.g., type of GPUs, internal cluster, or cloud provider). Yes. See Appendix
  \end{enumerate}

  \item If you are using existing assets (e.g., code, data, models) or curating/releasing new assets, check if you include:
  \begin{enumerate}
    \item Citations of the creator If your work uses existing assets. Yes
    \item The license information of the assets, if applicable. Yes
    \item New assets either in the supplemental material or as a URL, if applicable. No
    \item Information about consent from data providers/curators. Not Applicable
    \item Discussion of sensible content if applicable, e.g., personally identifiable information or offensive content. Not Applicable
  \end{enumerate}

  \item If you used crowdsourcing or conducted research with human subjects, check if you include:
  \begin{enumerate}
    \item The full text of instructions given to participants and screenshots. Not Applicable
    \item Descriptions of potential participant risks, with links to Institutional Review Board (IRB) approvals if applicable. Not Applicable
    \item The estimated hourly wage paid to participants and the total amount spent on participant compensation. Not Applicable
  \end{enumerate}

\end{enumerate}

\clearpage

\appendix
\thispagestyle{empty}

\onecolumn
\aistatstitle{Supplementary Materials \\ Hellinger Multimodal Variational Autoencoders}



\textbf{\large Table of Contents}
\startcontents[sections]
\printcontents[sections]{l}{1}{\setcounter{tocdepth}{2}}

The supplementary material is organized as follows: In Section~\ref{proof:lem_hel}, we provide the derivation of Hellinger aggregation and ablation studies on different types of experts across multiple aggregation methods. Section~\ref{app:exp} provides details on the dataset, evaluation criteria, as well as the experimental setup, including architecture and hyperparameters. Additional qualitative and quantitative results on three benchmark datasets are also included in Section~\ref{app:exp}, which could not be included in the main paper.

\section{AGGREGATION METHODS}

\label{proof:lem_hel}

\subsection{Derivation of Hellinger aggregation}

Consider the H\"older pooling function with $\alpha = 0.5$, applied to the set of unimodal posteriors $\{q_{\phi_j}(\vz|\vx_j)\}_{j=1}^M$. Each posterior is a multivariate Gaussian distributions with diagonal covariance matrix, $q_{\phi_j}(\vz|\vx_j) = \mathcal{N}\left(\vmu_j, \mathrm{diag}(\vsigma_j^2)\right)$, where $\vz = (z_1, z_2, \dots, z_D)^\top \in \sR^D$ denotes the latent variable, $\vmu_j = (\mu_{j,1}, \mu_{j,2}, \dots, \mu_{j,D})^\top \in \sR^D$ the mean vector, and $\mathrm{diag}(\vsigma_j^2) = \mathrm{diag}(\sigma_{j,1}^2, \sigma_{j,2}^2, \dots, \sigma_{j,D}^2)^\top \in \sR^{D \times D}$ the covariance matrix. For simplicity, we denote the unimodal posterior by $q_j(\vz)=q_{\phi_j}(\vz|\vx_j)$, and the approximate joint posterior by $q(\vz)= q_\phi(\vz|\sX)$. The pooled density is then defined as:
\begin{equation}
    q(\vz) = c \left(\sum_{\subindex=1}^\modalities \lambda_\subindex \sqrt{q_\subindex(\vz)}\right)^2,
\label{eqn:hel_pooling}
\end{equation}
where $\textstyle c = 1 / \int \left( \sum_{\subindex=1}^\modalities \lambda_\subindex \sqrt{q_\subindex(\vz)} \right)^2 d\vz$. As the exact pooling function is not Gaussian, we will project the optimal $q(\vz)$ in Equation~\ref{eqn:hel_pooling} back onto Gaussian family by moment matching, with mean and variance are defined as follow:
\begin{equation}
    \tilde{\vmu} = \int \vz q(\vz)d\vz = \frac{\int \vz \left(\sum_{j=1}^M \lambda_j \sqrt{q_j(\vz)}\right)^2 d\vz}{\int \left(\sum_{j=1}^M \lambda_j \sqrt{q_j(\vz)}\right)^2 d\vz},
\label{eqn:appox_mean}
\end{equation}
\begin{equation}
    \tilde{\mSigma} = \int \vz \vz^\top q(\vz)d\vz - \tilde{\vmu} \tilde{\vmu}^\top = \frac{\int \vz \vz^\top \left(\sum_{j=1}^M \lambda_j \sqrt{q_j(\vz)}\right)^2 d\vz}{\int \left(\sum_{j=1}^M \lambda_j \sqrt{q_j(\vz)}\right)^2 d\vz} - \tilde{\vmu} \tilde{\vmu}^\top.
\label{eqn:appox_variance}
\end{equation}
\textbf{Step 1.} We calculate the denominator term in both Equations~\ref{eqn:appox_mean}, and~\ref{eqn:appox_variance} for mean and variance:
\begin{align}
    \int \left(\sum_{j=1}^M \lambda_j \sqrt{q_j(\vz)}\right)^2 d\vz
    &= \sum_{j=1}^M \lambda_j^2 \int q_j(\vz) d\vz + 2 \sum_{i=1}^M \sum_{j>i}^M \lambda_i \lambda_j \int \sqrt{q_i(\vz) q_j(\vz)} d\vz \nonumber \\
    &= \sum_{j=1}^M \lambda_j^2 + 2 \sum_{i=1}^M \sum_{j>i}^M \lambda_i \lambda_j S_{ij},
\label{eqn:deno}
\end{align}
as $\int q_j(\vz) d\vz = 1$ and $S_{ij} = \int \sqrt{q_i(\vz) q_j(\vz)} d\vz$ is a Hellinger affinity between two Gaussians. As $q_i(\vz)$ and $q_j(\vz)$ are two multivariate Gaussian distributions with diagonal covariance matrix, they can be factorized across dimensions as follows:
\begin{equation*}
    q_i(\vz) = \prod_{d=1}^D q_{i,d}(z_d), \quad q_j(\vz) = \prod_{d=1}^D q_{j,d}(z_d),
\end{equation*}
where each $q_{i,d}(z_d)$ and $q_{j,d}(z_d)$ denotes a one-dimensional Gaussian distribution along the coordinate $z_d$. Hence, we can express $S_{ij}$ as:
\begin{equation}
    S_{ij} = \int \sqrt{q_i(\vz) q_j(\vz)} d\vz = \int \sqrt{\prod_{d=1}^D q_{i,d}(z_d) q_{j,d}(z_d)} d\vz = \prod_{d=1}^D \int \sqrt{q_{i,d}(z_d) q_{j,d}(z_d)} dz_d.
\label{eqn:Sij_complex}
\end{equation}
To derive further, we need to know the density of $q_{i,d}(z_d)$ and $q_{j,d}(z_d)$:
\begin{equation*}
    q_{i,d}(z_d) = \frac{1}{\sqrt{2\pi} \sigma_{i,d}} \exp{\left(-\frac{(z_d - \mu_{i,d})^2}{2\sigma_{i,d}^2}\right)}, \quad q_{j,d}(z_d) = \frac{1}{\sqrt{2\pi} \sigma_{j,d}} \exp{\left(-\frac{(z_d - \mu_{j,d})^2}{2\sigma_{j,d}^2}\right)}.
\end{equation*}
Then, $\int \sqrt{q_{i,d}(z_d) q_{j,d}(z_d)} dz_d$ becomes:
\begin{equation*}
    \int \sqrt{q_{i,d}(z_d) q_{j,d}(z_d)} dz_d = \int \frac{1}{\sqrt{2\pi \sigma_{i,d} \sigma_{j,d}}} \exp{\left(-\frac14 \left[ \frac{(z_d-\mu_{i,d})^2}{\sigma_{i,d}^2} + \frac{(z_d-\mu_{j,d})^2}{\sigma_{j,d}^2} \right] \right)} dz_d.
\end{equation*}
Next, we expand the expression inside the exponential term:
\begin{align*}
    -\frac14 \left[ \frac{(z_d-\mu_{i,d})^2}{\sigma_{i,d}^2} + \frac{(z_d-\mu_{j,d})^2}{\sigma_{j,d}^2} \right] &= - \frac12 A_dz_d^2 + B_dz_d + C_d \\
    &= - \frac12 A_d \left[z_d^2 - 2 \frac{B_d}{A_d} z_d + \left(\frac{B_d}{A_d}\right)^2\right] + \frac{B_d^2}{2A_d} + C_d \\
    &= -\frac12 A_d \left(z_d - \frac{B_d}{A_d}\right)^2 + \frac{B_d^2}{2A_d} + C_d,
\end{align*}
where
\begin{equation}
    A_d = \frac12 \left(\frac{1}{\sigma_{i,d}^2} + \frac{1}{\sigma_{j,d}^2}\right), \quad B_d = \frac12 \left( \frac{\mu_{i,d}}{\sigma_{i,d}^2} + \frac{\mu_{j,d}}{\sigma_{j,d}^2}\right), \quad C_d = -\frac14 \left( \frac{\mu_{i,d}^2}{\sigma_{i,d}^2} + \frac{\mu_{j,d}^2}{\sigma_{j,d}^2}\right).
\label{eqn:ABC}
\end{equation}
Combining all the above terms, we obtain an expression proportional to a Gaussian in $z_d$. Consequently, we can write $\int \sqrt{q_{i,d}(z_d) q_{j,d}(z_d)} dz_d$ as:
\begin{align}
    \int \sqrt{q_{i,d}(z_d) q_{j,d}(z_d)} dz_d &= \int \frac{1}{\sqrt{2\pi \sigma_{i,d} \sigma_{j,d}}} \exp{\left( -\frac12 A_d \left(z_d - \frac{B_d}{A_d}\right)^2 + \frac{B_d^2}{2A_d} + C_d \right)} dz_d \nonumber \\
    &= \int \frac{1}{\sqrt{2\pi \sigma_{i,d} \sigma_{j,d}}} \exp{\left(\frac{B_d^2}{2A_d} + C_d\right)} \exp{\left(-\frac{(z_d - \frac{B_d}{A_d})^2}{2A_d^{-1}}\right)} dz_d \nonumber \\
    &= \frac{A_d^{-1/2}}{\sqrt{\sigma_{i,d} \sigma_{j,d}}} \exp{\left(\frac{B_d^2}{2A_d} + C_d\right)}.
\label{eqn:hel_affinity_1}
\end{align}
The first term in the product is given by:
\begin{equation*}
    \frac{A_d^{-1/2}}{\sqrt{\sigma_{i,d} \sigma_{j,d}}} 
    = \frac{1}{\sqrt{\sigma_{i,d}\sigma_{j,d}}} \left(\frac{\sigma_{i,d}^2+\sigma_{j,d}^2}{2\sigma_{i,d}^2\sigma_{j,d}^2}\right)^{-1/2} = \sqrt{\frac{2\sigma_{i,d}\sigma_{j,d}}{\sigma_{i,d}^2+\sigma_{j,d}^2}}.
\end{equation*}
The second term in the product is given by:
\begin{equation*}
    \exp{\left(\frac{B_d^2}{2A_d} + C_d\right)} = \exp{\left(-\frac14 \frac{(\mu_{i,d}-\mu_{j,d})^2}{\sigma_{i,d}^2+\sigma_{j,d}^2}\right)}.
\end{equation*}
Then, we obtain the final expression for $\int \sqrt{q_{i,d}(z_d) q_{j,d}(z_d)} dz_d$:
\begin{equation}
    \int \sqrt{q_{i,d}(z_d) q_{j,d}(z_d)} dz_d = \sqrt{\frac{2 \sigma_{i,d} \sigma_{j,d}}{\sigma_{i,d}^2 + \sigma_{j,d}^2}} \exp\left( -\frac{1}{4} \frac{(\mu_{i,d} - \mu_{j,d})^2}{\sigma_{i,d}^2 + \sigma_{j,d}^2} \right).
\label{eqn:hel_affinity_2}
\end{equation}
For each dimension $d$, this term corresponds to the Hellinger affinity between the one-dimensional Gaussians $q_{i,d}(z_d)$ and $q_{j,d}(z_d)$. Hence, $S_{ij}$ in Equation~\ref{eqn:Sij_complex} can be rewritten as:
\begin{equation}
    S_{ij} = \prod_{d=1}^D \sqrt{\frac{2 \sigma_{i,d} \sigma_{j,d}}{\sigma_{i,d}^2 + \sigma_{j,d}^2}} \exp\left( -\frac{1}{4} \frac{(\mu_{i,d} - \mu_{j,d})^2}{\sigma_{i,d}^2 + \sigma_{j,d}^2} \right).
\label{eqn:Sij_simple}
\end{equation}
\textbf{Step 2.} We compute the numerator in Equation~\ref{eqn:appox_mean} for the mean:
\begin{align}
    \int \vz \left(\sum_{j=1}^M \lambda_j \sqrt{q_j(\vz)}\right)^2 d\vz
    &= \sum_{j=1}^M \lambda_j^2 \int \vz q_j(\vz) d\vz + 2 \sum_{i=1}^M \sum_{j>i}^M \lambda_i \lambda_j \int \vz \sqrt{q_i(\vz) q_j(\vz)} d\vz \nonumber \\
    &= \sum_{j=1}^M \lambda_j^2 \vmu_j + 2 \sum_{i=1}^M \sum_{j>i}^M \lambda_i \lambda_j \mM_{ij},
\label{eqn:nomi_mean}
\end{align}
as $\int \vz q_j(\vz) d\vz = \vmu_j$ and $\mM_{ij} = \int \vz \sqrt{q_i(\vz) q_j(\vz)} d\vz$. Vector $\mM_{ij} \in \sR^D$ represents the first moment of $\sqrt{q_i(\vz) q_j(\vz)}$, it can be computed analogously to $S_{ij}$ by factorizing across dimensions. Denoting $\mM_{ij} = (M_{ij,1}, M_{ij,2}, \dots, M_{ij,D})^\top$, we obtain $M_{ij,d}$ for each coordinate $d$:
\begin{align*}
    M_{ij,d} &=  \int z_d \sqrt{q_i(\vz) q_j(\vz)} d\vz \\ 
    &= \left[\int z_d \sqrt{q_{i,d}(z_d) q_{j,d}(z_d)} dz_d\right] \prod_{m\neq d}^D\left[\int \sqrt{q_{i,m}(z_m) q_{j,m}(z_m)} dz_m\right].
\end{align*}
The first term in the product is computed in the same way as in Equations~\ref{eqn:hel_affinity_1} and \ref{eqn:hel_affinity_2}:
\begin{align}
    \int z_d \sqrt{q_{i,d}(z_d), q_{j,d}(z_d)} dz_d 
    &= \int \frac{1}{\sqrt{2\pi \sigma_{i,d} \sigma_{j,d}}} \exp{\left(\frac{B_d^2}{2A_d} + C_d\right)} z_d \exp{\left(-\frac{(z_d - \frac{B_d}{A_d})^2}{2A_d^{-1}}\right)} dz_d \nonumber \\
    &= \frac{A_d^{-1/2}}{\sqrt{\sigma_{i,d} \sigma_{j,d}}} \exp{\left(\frac{B_d^2}{2A_d} + C_d\right)} \frac{B_d}{A_d} \nonumber \\
    &= \sqrt{\frac{2 \sigma_{i,d} \sigma_{j,d}}{\sigma_{i,d}^2 + \sigma_{j,d}^2}} \exp\left( -\frac{1}{4} \frac{(\mu_{i,d} - \mu_{j,d})^2}{\sigma_{i,d}^2 + \sigma_{j,d}^2} \right) \mu_{ij,d},
\label{eqn:first_moment}
\end{align}
where
\begin{equation}
    \mu_{ij,d} = \frac{B_d}{A_d} = \frac{\mu_{i,d}\sigma_{j,d}^2+\mu_{j,d}\sigma_{i,d}^2}{\sigma_{i,d}^2 + \sigma_{j,d}^2},
\label{eqn:muijd}
\end{equation}
be derived from Equation~\ref{eqn:ABC}. The second term in the product can be computed from Equation~\ref{eqn:hel_affinity_2}:
\begin{equation*}
    \prod_{m\neq d}^D \int \sqrt{q_{i,m}(z_m) q_{j,m}(z_m)} dz_m = \prod_{m\neq d}^D \sqrt{\frac{2 \sigma_{i,m} \sigma_{j,m}}{\sigma_{i,m}^2 + \sigma_{j,m}^2}} \exp\left( -\frac{1}{4} \frac{(\mu_{i,m} - \mu_{j,m})^2}{\sigma_{i,m}^2 + \sigma_{j,m}^2} \right) 
\end{equation*}
Then, we obtain the final expression for $M_{ij,d}$:
\begin{equation}
    M_{ij,d} = \mu_{ij,d} \prod_{d=1}^D \sqrt{\frac{2 \sigma_{i,d} \sigma_{j,d}}{\sigma_{i,d}^2 + \sigma_{j,d}^2}} \exp\left( -\frac{1}{4} \frac{(\mu_{i,d} - \mu_{j,d})^2}{\sigma_{i,d}^2 + \sigma_{j,d}^2} \right) = \mu_{ij,d} S_{ij}.
\label{eqn:Mijd}
\end{equation}
Therefore, $\mM_{ij} = (\mu_{ij,1}, \mu_{ij, 2}, \dots, \mu_{ij, D})^\top S_{ij}.$

\textbf{Step 3.} The mean $\tilde{\vmu}$ is obtained by combining Equations~\ref{eqn:deno},~\ref{eqn:nomi_mean}, and~\ref{eqn:Mijd}, yielding:
\begin{equation*}
    \tilde{\vmu} = \frac{\sum_{j=1}^M \lambda_j^2 \vmu_j + 2 \sum_{i=1}^M \sum_{j>i}^M \lambda_i \lambda_j \mM_{ij}}{ \sum_{j=1}^M \lambda_j^2 + 2 \sum_{i=1}^M \sum_{j>i}^M \lambda_i \lambda_j S_{ij}}.
\end{equation*}
For each dimension $d$, the component $\tilde{\mu}_d$ of $\tilde{\vmu} = (\tilde{\mu}_1, \tilde{\mu}_2, \dots, \tilde{\mu}_D) \in \sR^D$ can be expressed as:
\begin{equation}
    \tilde{\mu}_d = \frac{\sum_{j=1}^M \lambda_j^2 \mu_{j,d} + 2 \sum_{i=1}^M \sum_{j>i}^M \lambda_i \lambda_j M_{ij, d}}{ \sum_{j=1}^M \lambda_j^2 + 2 \sum_{i=1}^M \sum_{j>i}^M \lambda_i \lambda_j S_{ij}} =  \frac{\sum_{j=1}^M \lambda_j^2 \mu_{j,d} + 2 \sum_{i=1}^M \sum_{j>i}^M \lambda_i \lambda_j \mu_{ij, d} S_{ij}}{ \sum_{j=1}^M \lambda_j^2 + 2 \sum_{i=1}^M \sum_{j>i}^M \lambda_i \lambda_j S_{ij}}.
\label{eqn:tildemud}
\end{equation}
\textbf{Step 4.} We compute the numerator in Equation~\ref{eqn:appox_variance} for the variance:
\begin{align}
    \int_{z} \vz \vz^\top \left(\sum_{j=1}^M \lambda_j \sqrt{q_j(\vz)}\right)^2 d\vz
    &= \sum_{j=1}^M \lambda_j^2 \int_{z} \vz \vz^\top q_j(\vz) d\vz + 2 \sum_{i=1}^M \sum_{j>i}^M \lambda_i \lambda_j \int_{z} \vz \vz^\top \sqrt{q_i(\vz) q_j(\vz)} d\vz \nonumber \\
    &= \sum_{j=1}^M \lambda_j^2 (\vmu_j \vmu_j^\top + \mathrm{diag}(\vsigma_j^2)) + 2 \sum_{i=1}^M \sum_{j>i}^M \lambda_i \lambda_j \mV_{ij},
\label{eqn:nomi_variance}
\end{align}
as $\int_{z} \vz \vz^\top q_j(\vz) d\vz = \vmu_j \vmu_j^\top + \mathrm{diag}(\vsigma_j^2)$ and $\mV_{ij} = \int \vz \vz^\top \sqrt{q_i(\vz) q_j(\vz)} d\vz$. Matrix $\mV_{ij}$ represents the second moment of $\sqrt{q_i(\vz) q_j(\vz)}$, it can be computed analogously to $S_{ij}$ by factorizing across dimensions. For the off diagonal term of $\mV_{ij}$ (when $d \neq e$), the expression factorizes as:
\begin{align*}
    V_{ij, d, e} &=  \int z_d z_e \sqrt{q_i(\vz) q_j(\vz)} d\vz \\ 
    &= \left[\int z_d \sqrt{q_{i,d}(z_d) q_{j,d}(z_d)} dz_d\right] \left[\int z_e \sqrt{q_{i,e}(z_e) q_{j,e}(z_e)} dz_e\right] \prod_{m\neq d, e}^D\left[\int \sqrt{q_{i,m}(z_m) q_{j,m}(z_m)} dz_m\right] \\
    &= \mu_{ij,d} \mu_{ij,e} \prod_{d=1}^D \sqrt{\frac{2 \sigma_{i,d} \sigma_{j,d}}{\sigma_{i,d}^2 + \sigma_{j,d}^2}} \exp\left( -\frac{1}{4} \frac{(\mu_{i,d} - \mu_{j,d})^2}{\sigma_{i,d}^2 + \sigma_{j,d}^2} \right)\\
    &= \mu_{ij,d} \mu_{ij,e} S_{ij},
\end{align*}
be derived from Equations~\ref{eqn:hel_affinity_2} and~\ref{eqn:first_moment}. For the diagonal term of $\mV_{ij}$ (when $d = e$), the expression factorizes as:
\begin{align*}
    V_{ij, d, d} &=  \int z_d^2 \sqrt{q_i(\vz) q_j(\vz)} d\vz \\ 
    &= \left[\int z_d^2 \sqrt{q_{i,d}(z_d) q_{j,d}(z_d)} dz_d\right] \prod_{m\neq d}^D\left[\int \sqrt{q_{i,m}(z_m) q_{j,m}(z_m)} dz_m\right].
\end{align*}
The first term in the product is computed in the same way as in Equations~\ref{eqn:hel_affinity_1} and \ref{eqn:hel_affinity_2}:
\begin{align*}
    \int z_d^2 \sqrt{q_{i,d}(z_d), q_{j,d}(z_d)} dz_d 
    &= \int \frac{1}{\sqrt{2\pi \sigma_{i,d} \sigma_{j,d}}} \exp{\left(\frac{B_d^2}{2A_d} + C_d\right)} z_d^2 \exp{\left(-\frac{(z_d - \frac{B_d}{A_d})^2}{2A_d^{-1}}\right)} dz_d \\
    &= \frac{A_d^{-1/2}}{\sqrt{\sigma_{i,d} \sigma_{j,d}}} \exp{\left(\frac{B_d^2}{2A_d} + C_d\right)} \left(\frac{B_d^2}{A_d^2} + A_d^{-1}\right) \\
    &= \sqrt{\frac{2 \sigma_{i,d} \sigma_{j,d}}{\sigma_{i,d}^2 + \sigma_{j,d}^2}} \exp\left( -\frac{1}{4} \frac{(\mu_{i,d} - \mu_{j,d})^2}{\sigma_{i,d}^2 + \sigma_{j,d}^2} \right) (\mu_{ij,d}^2 + \sigma_{ij,d}^2),
\end{align*}
where
\begin{equation}
    \sigma_{ij,d}^2 = A_d^{-1} = \frac{2\sigma_{i,d}^2\sigma_{j,d}^2}{\sigma_{i,d}^2+\sigma_{j,d}^2},
\label{eqn:sigmaijd}
\end{equation}
be derived from Equation~\ref{eqn:ABC}. The second term in the product follows from Equation~\ref{eqn:hel_affinity_2}. Thus, $V_{ij,d,d}$ becomes:
\begin{equation}
    V_{ij,d,d} = (\mu_{ij,d}^2 + \sigma_{ij,d}^2) \prod_{d=1}^D \sqrt{\frac{2 \sigma_{i,d} \sigma_{j,d}}{\sigma_{i,d}^2 + \sigma_{j,d}^2}} \exp\left( -\frac{1}{4} \frac{(\mu_{i,d} - \mu_{j,d})^2}{\sigma_{i,d}^2 + \sigma_{j,d}^2} \right) = (\mu_{ij,d}^2 + \sigma_{ij,d}^2) S_{ij}.
\label{eqn:Vijdd}
\end{equation}
Therefore, we can express matrix $\mV_{ij}$ by using off diagonal terms $V_{ij,d,e}= \mu_{ij,d} \mu_{ij,e} S_{ij}$ when $d\neq e$ and diagonal term $V_{ij,d,d}=(\mu_{ij,d}^2 + \sigma_{ij,d}^2) S_{ij}$ when $d = e$.

\textbf{Step 5.} The variance $\tilde{\mSigma}$ is obtained by combining Equations~\ref{eqn:deno},~\ref{eqn:nomi_variance}, and~\ref{eqn:Vijdd}, yielding:
\begin{equation*}
    \tilde{\mSigma} = \frac{\sum_{j=1}^M \lambda_j^2 (\vmu_j \vmu_j^\top + \vsigma_j^2 \mI_D) + 2 \sum_{i=1}^M \sum_{j>i}^M \lambda_i \lambda_j \mV_{ij}}{ \sum_{j=1}^M \lambda_j^2 + 2 \sum_{i=1}^M \sum_{j>i}^M \lambda_i \lambda_j S_{ij}} - \tilde{\vmu} \tilde{\vmu}^\top.
\end{equation*}
By approximating $q(\vz)$ with a diagonal covariance matrix, 
we simplify $\tilde{\mSigma}$ by removing the off-diagonal terms, yielding 
$\tilde{\mSigma} \approx \mathrm{diag}(\tilde{\sigma}_1^2, \tilde{\sigma}_2^2, \dots, \tilde{\sigma}_D^2)$, 
where each $\tilde{\sigma}_d^2$ corresponds to dimension $d$ and can be expressed as:
\begin{align}
    \tilde{\sigma}_d^2 &= \frac{\sum_{j=1}^M \lambda_j^2 (\mu_{j,d}^2 + \sigma_{j,d}^2) + 2 \sum_{i=1}^M \sum_{j>i}^M \lambda_i \lambda_j V_{ij, d, d}}{ \sum_{j=1}^M \lambda_j^2 + 2 \sum_{i=1}^M \sum_{j>i}^M \lambda_i \lambda_j S_{ij}} - (\tilde{\mu}_d)^2 \nonumber \\ 
    &= \frac{\sum_{j=1}^M \lambda_j^2 (\mu_{j,d}^2 + \sigma_{j,d}^2) + 2 \sum_{i=1}^M \sum_{j>i}^M \lambda_i \lambda_j (\mu_{ij,d}^2 + \sigma_{ij,d}^2) S_{ij}}{ \sum_{j=1}^M \lambda_j^2 + 2 \sum_{i=1}^M \sum_{j>i}^M \lambda_i \lambda_j S_{ij}} - (\tilde{\mu}_d)^2.
\label{eqn:tildesigmad}
\end{align}
\textbf{Step 6.} Assuming $\lambda_j = 1/M$ for all $j$, we approximate $q(\vz) \sim \tilde{q}(\vz) = \mathcal{N}(\tilde{\vmu}, \mathrm{diag}(\tilde{\vsigma}^2))$, with $\tilde{\vmu} = (\tilde{\mu}_{1}, \tilde{\mu}_2 \dots, \tilde{\mu}_{D})^\top \in \sR^D$ and $\tilde{\vsigma}^2 = (\tilde{\sigma}_{1}^2, \tilde{\sigma}_2^2 \dots, \tilde{\sigma}_{D}^2)^\top \in \sR^D$. For each dimension $d$, Equations~\ref{eqn:tildemud} and \ref{eqn:tildesigmad} give:
\begin{equation*}
    \tilde{\mu}_d =  \frac{\sum_{j=1}^M \mu_{j,d} + 2 \sum_{i=1}^M \sum_{j>i}^M \mu_{ij, d} S_{ij}}{ M+ 2 \sum_{i=1}^M \sum_{j>i}^M S_{ij}},
\end{equation*}
\begin{equation*}
    \tilde{\sigma}_d^2 = \frac{\sum_j (\mu_{j,d}^2 + \sigma_{j,d}^2) + 2 \sum_{i=1}^M \sum_{j>i}^M (\mu_{ij, d}^2 + \sigma_{ij, d}^2) S_{ij}}{\modalities + 2 \sum_{i=1}^M \sum_{j>i}^M S_{ij}} - (\tilde{\mu}_d)^2,
\end{equation*}
with $\mu_{ij,d}$, $\sigma_{ij,d}$, and $S_{ij}$ as defined in Equations~\ref{eqn:muijd}, \ref{eqn:sigmaijd}, and \ref{eqn:Sij_simple}.  \hfill $\square$

\subsection{Hellinger aggregation beyond Gaussians: the exponential family}

Our proposed Hellinger aggregation is not restricted to Gaussians: it can be adapted to many distributions in the exponential family for which the closed-form expressions of their Hellinger distances are known. Assume each modality posterior $q_j(\vz)$ belongs to the same exponential family with a common base measure $h(\vz)$, a fixed sufficient statistics $T(\vz)$, and common support $\mathcal{Z}$: $q_\eta(\vz) = h(\vz)\exp(\eta^\top T(\vz)-A(\eta)), \vz \in \mathcal{Z}$. Let modality $j$ have natural parameter $\eta_j$ for all $j \in \{1, \dots, M\}$, we then expand the cross-term $\sqrt{q_i(\vz)q_j(\vz)}$.
\begin{align*}
    \sqrt{q_i(\vz)q_j(\vz)} &= \sqrt{q_{\eta_i}(\vz)q_{\eta_j}(\vz)} \\
    &= \sqrt{h(\vz)\exp(\eta_i^\top T(\vz)-A(\eta_i)) h(\vz) \exp(\eta_j^\top T(\vz)-A(\eta_i))} \\
    &= h(\vz) \exp \left(\frac{\eta_i^\top + \eta_j^\top}{2} T(\vz) - \frac{A(\eta_i) + A(\eta_j)}{2}\right)
    \\
    &= h(\vz) \exp \left(\eta_{ij}^\top T(\vz) - \frac{A(\eta_i) + A(\eta_j)}{2}\right), \quad \eta_{ij} = \frac{\eta_i + \eta_j}{2}.
\end{align*}
Hence, we can expand the Hellinger affinity $S_{ij} = \int \sqrt{q_i(\vz) q_j(\vz)} d\vz$ as:
\begin{align*}
    S_{ij} &= \int h(\vz) \exp \left(\eta_{ij}^\top T(\vz) - \frac{A(\eta_i) + A(\eta_j)}{2}\right) d\vz \\
    &= \exp \left(-\frac{A(\eta_i) + A(\eta_j)}{2}\right) \int h(\vz)\exp(\eta_{ij}^\top T(\vz)) d\vz \\
    &= \exp \left(-\frac{A(\eta_i) + A(\eta_j)}{2}\right) \exp(A(\eta_{ij})) \\
    &= \exp \left(A(\eta_{ij}) - \frac12 A(\eta_i) - \frac12 A(\eta_j)\right).
\end{align*}
Next, we will derive the first moment of $\sqrt{q_i(\vz) q_j(\vz)}$, namely $\mM_{ij} = \int \vz \sqrt{q_i(\vz) q_j(\vz)} d\vz$.
\begin{align*}
    \mM_{ij} &= \int \vz h(\vz) \exp \left(\eta_{ij}^\top T(\vz) - \frac{A(\eta_i) + A(\eta_j)}{2}\right) d\vz  \\
    &= \exp \left(-\frac{A(\eta_i) + A(\eta_j)}{2}\right) \int \vz h(\vz)\exp(\eta_{ij}^\top T(\vz)) d\vz \\
    &= \exp \left(A(\eta_{ij}) - \frac12 A(\eta_i) - \frac12 A(\eta_j)\right) \int \vz h(\vz)\exp(\eta_{ij}^\top T(\vz) - A(\eta_{ij})) d\vz \\
    &= S_{ij} \int z q_{\eta_{ij}}(\vz) d\vz \\
    &= S_{ij} \mathbb{E}_{q_{\eta_{ij}}}[\vz].
\end{align*}

Similarly, for the second moment of $\sqrt{q_i(\vz) q_j(\vz)}$, we can derive $\mV_{ij} = \int \vz \vz^\top \sqrt{q_i(\vz) q_j(\vz)} d\vz = S_{ij} \mathbb{E}_{q_{\eta_{ij}}}[\vz \vz^\top]$. 

Since exponential-family distributions satisfy $\mathbb{E}_{q_\eta}[T(\vz)] = \nabla A(\eta)$ and $\text{Cov}_{q_\eta}[T(\vz)] = \nabla^2 A(\eta)$, if $\vz$ and $\vz \vz^\top$ are included in $T(\vz)$ (e.g., for a Gaussian), this yields $\mathbb{E}_{q_{\eta_{ij}}}[\vz]$ and  $\mathbb{E}_{q_{\eta_{ij}}}[\vz \vz^\top]$ immediately; otherwise these moments can be obtained from the known closed-form moments of the chosen family evaluated at $\eta_{ij}$. 

\paragraph{Example: Exponential distribution.} Let each unimodal distribution be $q_j(\vz) = \text{Exp} (\alpha_j) = \alpha_j e^{-\alpha_j \vz}$ on $\vz \geq 0$. The corresponding exponential-family components are: 
\begin{equation*}
    T(\vz) = \vz, \quad h(\vz) = \mathbf{1}\{\vz \geq 0\}, \quad \eta_j = -\alpha_j, \quad A(\eta_j) = -\log(-\eta_j).
\end{equation*}
Using these, we can calculate the Hellinger affinity $S_{ij}$ as:
\begin{equation*}
    \eta_{ij} = -\frac{\alpha_i + \alpha_j}{2}, \quad S_{ij} = \exp \left(-\log\left(\frac{\alpha_i + \alpha_j}{2}\right) + \frac12 \log \alpha_i + \frac12 \log \alpha_j\right) = \frac{2\sqrt{\alpha_i \alpha_j}}{\alpha_i + \alpha_j}.
\end{equation*}
Since $q_{\eta_{ij}}(\vz) = \text{Exp}\left(\frac{\alpha_i + \alpha_j}{2}\right)$, we have
\begin{equation*}
    \mathbb{E}_{q_{\eta_{ij}}}[\vz] = \frac{2}{\alpha_i + \alpha_j}, \quad \mathbb{E}_{q_{\eta_{ij}}}[\vz \vz^\top] = \frac{8}{(\alpha_i + \alpha_j)^2}.
\end{equation*}
Therefore, we can obtain $M_{ij}$ and $V_{ij}$.

\subsection{Effect of the number of good and bad experts on multiple aggregation methods}

\begin{figure}[h]
    \begin{subfigure}{0.5\textwidth}
        \centering
        \includegraphics[width=0.98\linewidth]{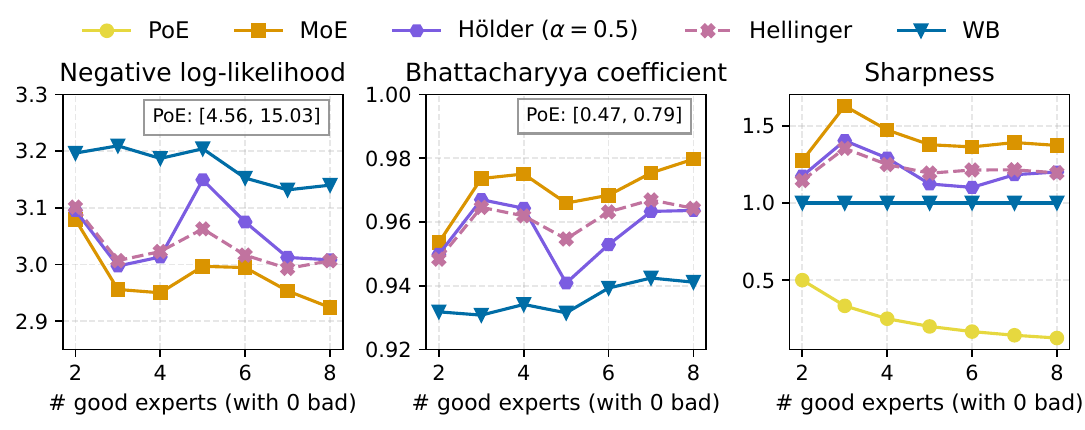}
    \end{subfigure}
    \hfill
    \begin{subfigure}{0.5\textwidth}
        \centering
        \includegraphics[width=0.98\linewidth]{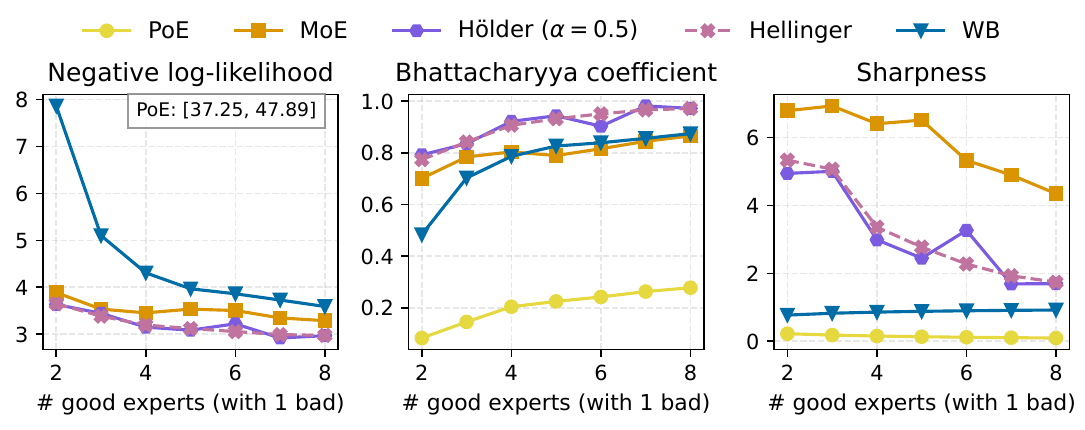}
    \end{subfigure}
    \vspace{-0.1in} 
    
    \begin{subfigure}{0.5\textwidth}
        \centering
        \includegraphics[width=0.98\linewidth]{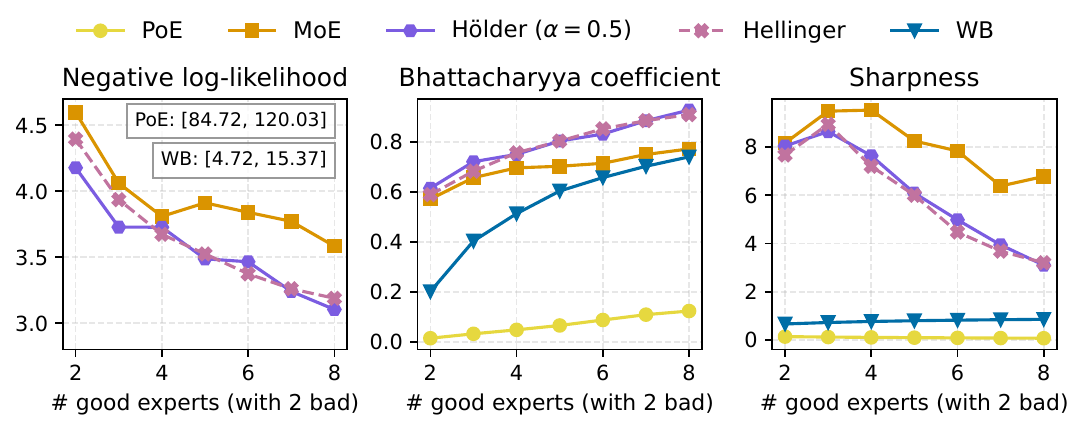}
    \end{subfigure}
    \hfill
    \begin{subfigure}{0.5\textwidth}
        \centering
        \includegraphics[width=0.98\linewidth]{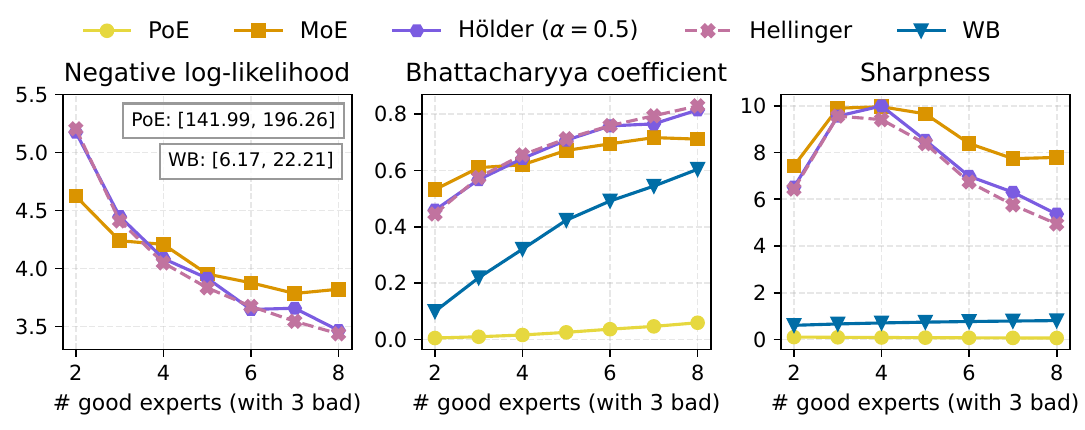}
    \end{subfigure}
    \vspace{-0.2in}
    \caption{Performance of PoE, MoE, Hölder pooling ($\alpha = 0.5$), Hellinger, and WB aggregators as a function of the number of good experts (with 0, 1, 2, and 3 bad experts fixed). Evaluation is based on negative log-likelihood ($\downarrow$), Bhattacharyya coefficient ($\uparrow$), and sharpness ($\downarrow$). Minimum and maximum NLL and BC values for PoE and WB are shown.}
    \label{fig:good_bad_expert_settings}
    \vspace{-0.1in}
\end{figure}

We extend the setting in Figure~\ref{fig:good_experts} by comparing with the Wasserstein barycenter (WB) from \citet{qiu2025multimodal}, which uses the 2-Wasserstein distance. As shown in Figure~\ref{fig:good_bad_expert_settings}, PoE suffers from overconfidence, resulting in significantly higher NLL and lower Bhattacharyya coefficients. WB is more stable than PoE, achieving a sharp distribution and reasonable overlap with the true distribution, though its NLL increases noticeably with more bad experts. Without bad experts, MoE performs slightly better than Hölder ($\alpha = 0.5$) and Hellinger, but with more bad experts, it yields higher NLL, lower Bhattacharyya coefficients, and less sharpness. This is because MoE and WB do not model dependencies between experts, which prevents them from distinguishing bad experts from good ones as their number increases. In contrast, for Hölder ($\alpha = 0.5$) and Hellinger, the overlap term $S_{ij}$ between good and bad experts is small, which lowers the contribution of bad experts to the joint posterior. As a result, they achieve the best balance across all metrics, showing strong robustness to different expert types.

\subsection{Effect of learnable pooling weights on model performance}


Since we use uniform pooling weights, we also trained HELVAE with learnable pooling weights to examine the difference. In Table~\ref{tab:helvae_pooling_weights}, on PolyMNIST at $\beta = 5$ (the best trade-off setting), uniform and learned pooling weights give comparable results, and the learned weights converge to nearly uniform values across modalities. In contrast, on CUB and CelebA, learning pooling weights degrades generative coherence and quality, as the optimization downweights noisier modalities in the aggregated posterior, allowing a few modalities to dominate. We therefore keep uniform pooling weights, as in PoE and MoE, since they provide a robust trade-off across modalities.

\begin{table}[h]
\caption{Generative quality and generative coherence on the PolyMNIST dataset. The table shows the results for HELVAE with uniform vs.\ learned pooling weights at $\beta = 5$.}
\vspace{-0.1in}
\label{tab:helvae_pooling_weights}
\begin{center}
\begin{tabular}{ccccc}
\toprule
& \multicolumn{2}{c}{Conditional} & \multicolumn{2}{c}{Unconditional} \\
& Coherence ($\uparrow$) & FID ($\downarrow$) & Coherence ($\uparrow$) & FID ($\downarrow$) \\
\midrule
HELVAE (uniform)       & 0.887 {\tiny $\pm$ 0.006} &  132.35 {\tiny $\pm$ 0.92} & 0.399 {\tiny $\pm$ 0.006} &  116.27 {\tiny $\pm$ 1.19} \\
HELVAE (learned)       & 0.886 {\tiny $\pm$ 0.002} &  130.25 {\tiny $\pm$ 1.84} & 0.421 {\tiny $\pm$ 0.028} &  115.20 {\tiny $\pm$ 1.04} \\
\bottomrule
\end{tabular}
\end{center}
\vspace{-0.15in}
\end{table}



\section{EXPERIMENTAL DETAILS}
\label{app:exp}

\subsection{Datasets}

\paragraph{PolyMNIST.} The dataset was introduced by~\citet{sutter2021generalized} as an extension of MNIST with varied backgrounds. A random $28 \times 28$ crop from five images serves as the background, forming a five-modality dataset. Shared information is the digit label, while background and handwriting style represent modality-specific features. The training and test sets contain $60{,}000$ and $10{,}000$ images, respectively.

\paragraph{CUB Image-Captions.} The Caltech Birds dataset consists of $11{,}788$ natural images of birds, each paired with 10 fine-grained captions describing appearance characteristics. The training and testing sets contain $88{,}550$ and $29{,}330$ samples, respectively. It has been widely used in multimodal VAE research~\citep{shi2019variational, daunhawer2021limitations, palumbo2023mmvae+, mancisidor2025aggregation}. However,~\citet{shi2019variational} employed a simplified version in which bird images were replaced with ResNet embeddings.

\paragraph{Bimodal CelebA.} Each CelebA image~\citep{liu2015deep} is annotated with $40$ facial attributes. The bimodal CelebA dataset, introduced by~\citet{sutter2020multimodal}, extends the original dataset with a text modality describing each face using these attributes. In this setting, negative attributes are omitted rather than explicitly negated, making learning more challenging since attributes appear in varying positions within the text, introducing additional variability. The training and test sets contain $162{,}770$ and $19{,}867$ samples, respectively.

\begin{figure}[h]
\begin{center}
\includegraphics[width=\textwidth]{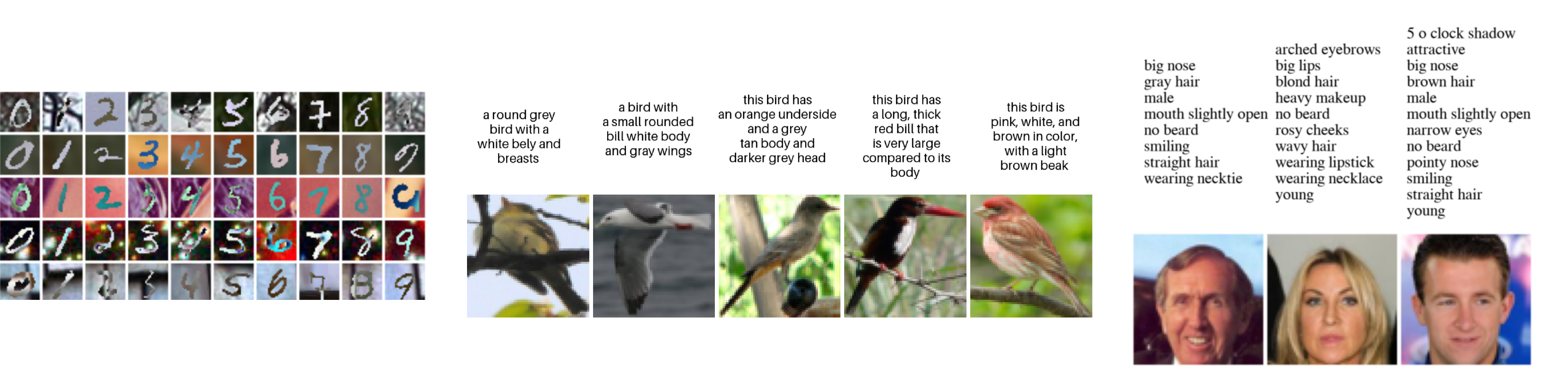}
\end{center}
\vspace{-0.2in}
\caption{ Illustrative samples from PolyMNIST (5 modalities), CUB Image-Captions (2 modalities) and bimodal CelebA (2 modalities), respectively.}
\label{fig:dataset}
\vspace{-0.1in}
\end{figure}

\subsection{Evaluation criteria}

\paragraph{Generative coherence (PolyMNIST and CelebA).} The coherence of conditionally generated samples reflects how well the imputed modalities align with the available ones in shared information. To compute generative coherence, we first train classifier networks on the training samples of each individual modality. Unconditional coherence is evaluated by classifying modalities generated from the prior and measuring the proportion assigned the same label. Conditional coherence is measured by classifying modalities generated from non-empty subsets that exclude the modality being evaluated. The reported generative coherence values are averaged over all generated images conditioned on all corresponding subsets and modalities.

\paragraph{Generative coherence (CUB).} We follow the approach of~\citet{palumbo2023mmvae+} to calculate conditional coherence. We construct captions of the form \textit{this bird is completely} [\textit{color}], where [\textit{color}] is one of \{\textit{white, yellow, red, blue, green, gray, brown, black}\}. For each caption, we generate ten images (eighty in total). We then count the number of pixels within the HSV ranges specified in Table~2 in~\citet{palumbo2023mmvae+}. An image is considered coherent if the caption color appears among the two dominant color classes (highest pixel counts). Conditional coherence is then reported as the fraction of coherent images over the total number generated.

\paragraph{Generative quality.} To evaluate generative quality, we use the Fréchet Inception Distance (FID) score~\citep{heusel2017gans}, a state-of-the-art metric for quantifying the visual quality of samples produced by generative models in image domains. FID measures the distance between the feature distributions of real and generated images, computed using activations of a pretrained Inception network, and has been shown to correlate well with human judgment. Lower FID values indicate closer alignment between generated and real data distributions.

\paragraph{Latent representation.} The quality of the learned representations serves as a proxy for their usefulness in downstream tasks beyond the training objective. High-quality representations of modality subsets are essential for conditionally generating coherent samples. We evaluate latent representations using logistic regression in scikit-learn~\citep{pedregosa2011scikit}. The classifier is trained on 500 encoded samples from the training set and evaluated on the test set. Reported results correspond to average performance across all test batches.

\subsection{Experimental setups}

We train all models using the reparameterization trick~\citep{kingma2013auto} and the Adam optimizer~\citep{kingma2014adam} on NVIDIA A100-PCIE-40GB GPUs with 64 CPU cores. Following prior work, KL terms in Equation~\ref{eqn:elbo} are weighted by a coefficient $\beta$~\citep{higgins2017beta}, i.e. $\beta \KL(q_\phi(\vz|\sX) \Vert p_\theta(\vz))$, selected via cross-validation from $\{1, 2.5, 5, 10\}$, with the best $\beta$ for each dataset reported in Table~\ref{tab:optimal_beta}. For all datasets, the prior is assumed to be an isotropic Gaussian, while unimodal posterior distributions are modeled as multivariate Gaussians with diagonal covariance. For all experiments, modality likelihoods are weighted by relative dimensionality, where the dominant modality is fixed to $1.0$ and the others are scaled proportionally to their data dimensions~\citep{shi2019variational, sutter2021generalized, javaloy2022mitigating}. We trained HELVAE with mixed precision. Results are reported as the mean and standard deviation across three independent runs. For benchmark models, we rely on the original implementations: \href{https://github.com/mhw32/multimodal-vae-public}{MVAE}, \href{https://github.com/iffsid/mmvae/tree/public}{MMVAE}, \href{https://github.com/thomassutter/MoPoE}{MoPoE}, \href{https://github.com/epalu/mmvaeplus/tree/new_release}{MMVAE+}, \href{https://github.com/rogelioamancisidor/codevae}{CoDEVAE} and open source library \href{https://github.com/AgatheSenellart/MultiVae}{MultiVAE}~\citep{Senellart2025}.

\begin{table}[h]
\caption{ Optimal $\beta$ for all models on the PolyMNIST, CUB and CelebA datasets.}
\vspace{-0.1in}
\label{tab:optimal_beta}
\begin{center}
\begin{tabular}{rccccccc}
\toprule
& MVAE & MMVAE & MoPoE & MMVAE+ & MWBVAE & CoDEVAE & HELVAE \\ \midrule
PolyMNIST & 2.5 & 2.5 & 5 & 2.5 & 2.5 & 5 & 5 \\
CUB  & 2.5 & 2.5 & 5 & 2.5 & 2.5 & 5 & 2.5\\
CelebA & 1 & 1 & 1 & 1 & 1 & 1 & 1\\
\bottomrule
\end{tabular}
\end{center}
\vspace{-0.1in}
\end{table}

\paragraph{PolyMNIST.} The encoder and decoder architectures follow~\citet{daunhawer2021limitations, palumbo2023mmvae+}, which employ ResNet backbones for all image modalities. We assume Laplace likelihoods for all modalities with Gaussian priors and posteriors, except for MMVAE and MMVAE+, where Laplace priors and posteriors are used. All models were trained for 500 epochs with an initial learning rate of $5e^{-4}$ and a latent space of size $512$, except for MMVAE and MMVAE+, which were trained for $150$ epochs. For MMVAE, the latent space size was set to $160$, while for MMVAE+ the shared and modality-specific latent subspaces were both set to $32$ dimensions, as suggested in the original implementations. The batch size was set to $256$.

\paragraph{CUB Image-Captions.} The encoder and decoder architectures follow~\citet{palumbo2023mmvae+}, using ResNet networks for image modality and CNNs for text modality. We assume Laplace and one-hot categorical likelihood distributions for images and captions, respectively, with Gaussian priors and posteriors. All models were trained with an initial learning rate of $5e^{-4}$ and a latent space of $64$ dimensions. For MMVAE+, the modality-specific latent space was set to $16$ dimensions and the shared latent space to $48$ dimensions. We trained MVAE, MoPoE, MWBVAE, CoDEVAE, and HELVAE for $150$ epochs, while MMVAE and MMVAE+ were trained for $50$ epochs with $K=10$, using the DReG estimator in the objective. The batch size for MMVAE and MMVAE+ was set to 32, while for the other models it was 128. 

\paragraph{Bimodal CelebA.} The encoder and decoder architectures follow~\citet{sutter2021generalized}, using residual blocks for both modalities. We assume Laplace and one-hot categorical likelihood distributions for images and captions, respectively, with Gaussian priors and posteriors. All models were trained with an initial learning rate of $1e^{-4}$ and a latent space of $32$ dimensions. Following~\citet{sutter2021generalized}, an additional modality-specific latent space of $32$ dimensions was added to each modality, resulting in a total latent dimension of $64$ per modality. The batch size was set to $256$, and models were trained for $200$ epochs.

\textit{\textbf{Note.} For MMVAE+ on the CUB dataset, we report results from the original source code and paper, as we were unable to reproduce the results in our experimental setting despite extensive efforts.}



\subsection{Additional results: PolyMNIST}
\label{app:pm}

\begin{figure}[h]
\begin{center}
\begin{subfigure}{\textwidth}
        \includegraphics[width=\linewidth]{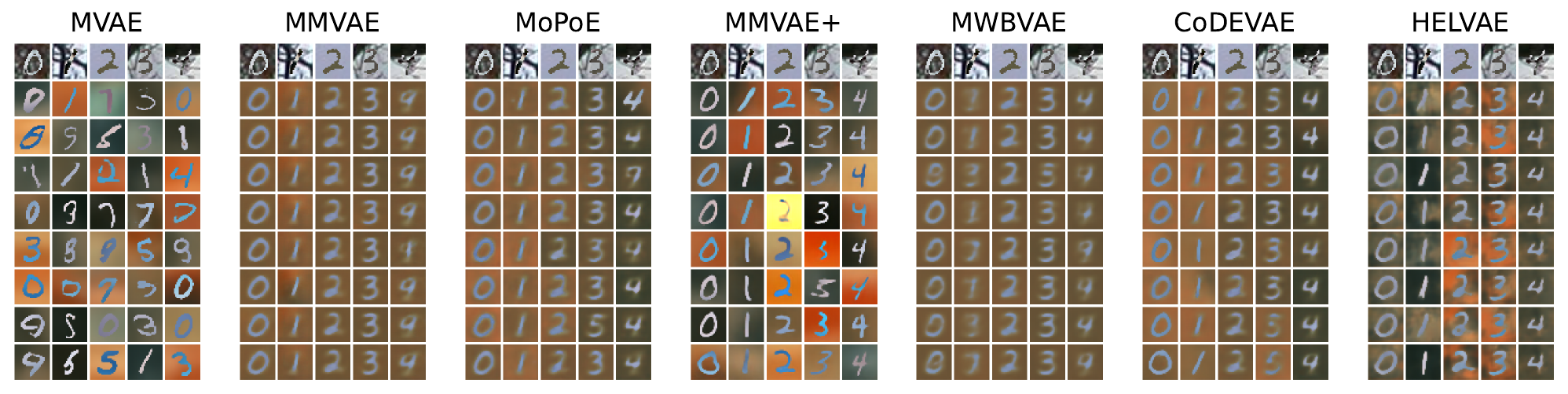}
    \end{subfigure}
    \vspace{0.5em} 
    \begin{subfigure}{\textwidth}
        \includegraphics[width=\linewidth]{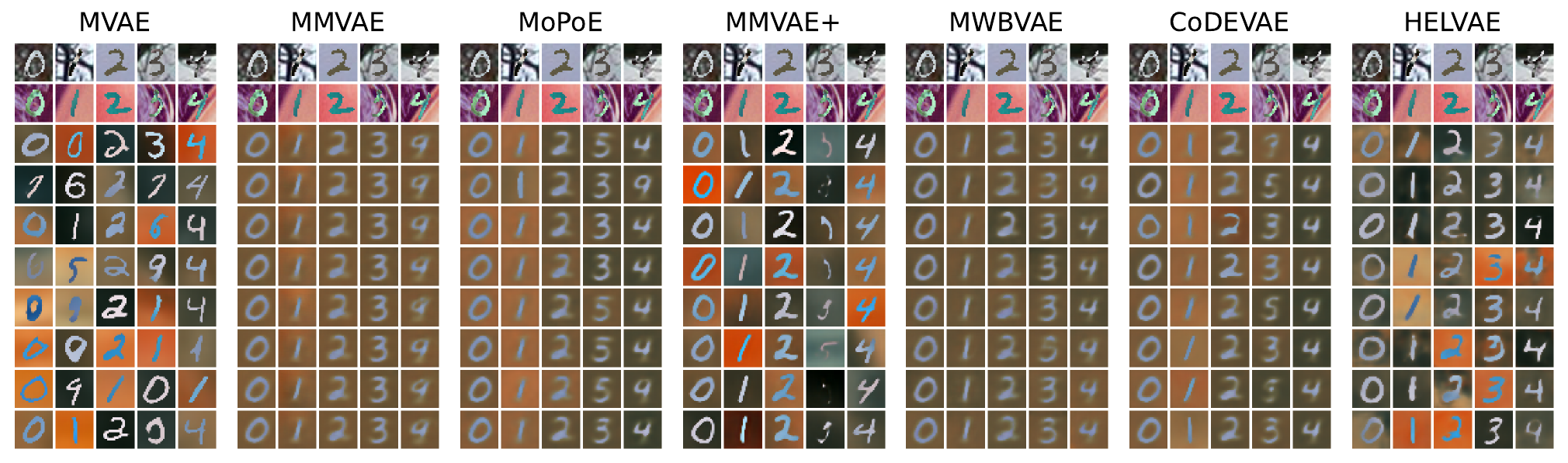}
    \end{subfigure}
    \vspace{0.5em} 
    \begin{subfigure}{\textwidth}
        \includegraphics[width=\linewidth]{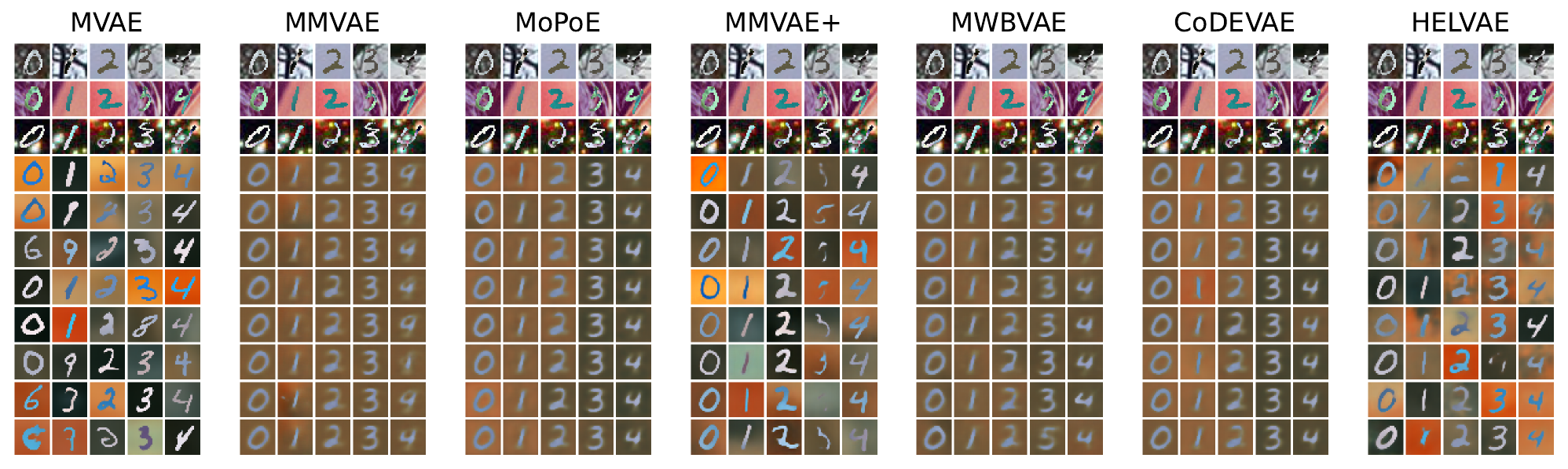}
    \end{subfigure}
    \vspace{0.5em} 
    \begin{subfigure}{\textwidth}
        \includegraphics[width=\linewidth]{plots/pm/pm_4_1.pdf}
    \end{subfigure}
\end{center}
\vspace{-0.1in}
\caption{ Conditionally generated samples of the second modality on the PolyMNIST dataset, with each model evaluated at its best $\beta \in \{1, 2.5, 5, 10\}$. The four panels show generations conditioned on (i) the first modality, (ii) the first and third, (iii) the first, third, and fourth, and (iv) the first, third, fourth, and fifth modalities. In each panel, the top rows show the conditioning inputs and the subsequent rows display the generated samples.}
\label{fig:pm_cond}
\end{figure}

In Table~\ref{tab:pm_all_results}, we report the numerical values corresponding to the quantitative comparison in Figure~\ref{fig:pm_coh_fid}. The results show that only HELVAE and MMVAE+ consistently achieve both high generative quality and high generative coherence across different $\beta$ values, for both conditional and unconditional generation. In contrast, alternative models either underperform on one of the two criteria or reach competitive scores only for conditional or unconditional generation (e.g.\ CoDEVAE). Furthermore, while MMVAE+ achieves its best performance at $\beta=2.5$, HELVAE outperforms MMVAE+ at both smaller $\beta=1$ and larger $\beta=10$ values.  Figure~\ref{fig:pm_cond} show qualitative results of conditionally generated samples of the second modality given corresponding test examples from one, two, three, and four modalities. Each column shows distinct samples drawn from the approximate joint posterior, ideally preserving digit identity while capturing stylistic variations. HELVAE generates more coherent samples when conditioned on four modalities compared to a single one, whereas MMVAE+ shows decreased coherence under the same setting, indicating that HELVAE effectively benefits from additional modalities.

\subsection{Additional results: CUB Image-Captions}
\label{app:cub}
\begin{figure}[h]
    \centering
    \begin{subfigure}{0.32\textwidth}
        \includegraphics[width=\linewidth]{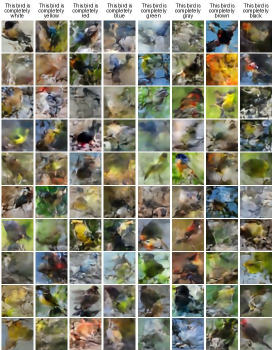}
        \caption{ MVAE}
    \end{subfigure}
    \hfill
    \begin{subfigure}{0.32\textwidth}
        \includegraphics[width=\linewidth]{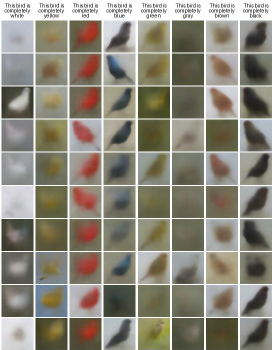}
        \caption{ MMVAE}
    \end{subfigure}
    \hfill
    \begin{subfigure}{0.32\textwidth}
        \includegraphics[width=0.9935\linewidth]{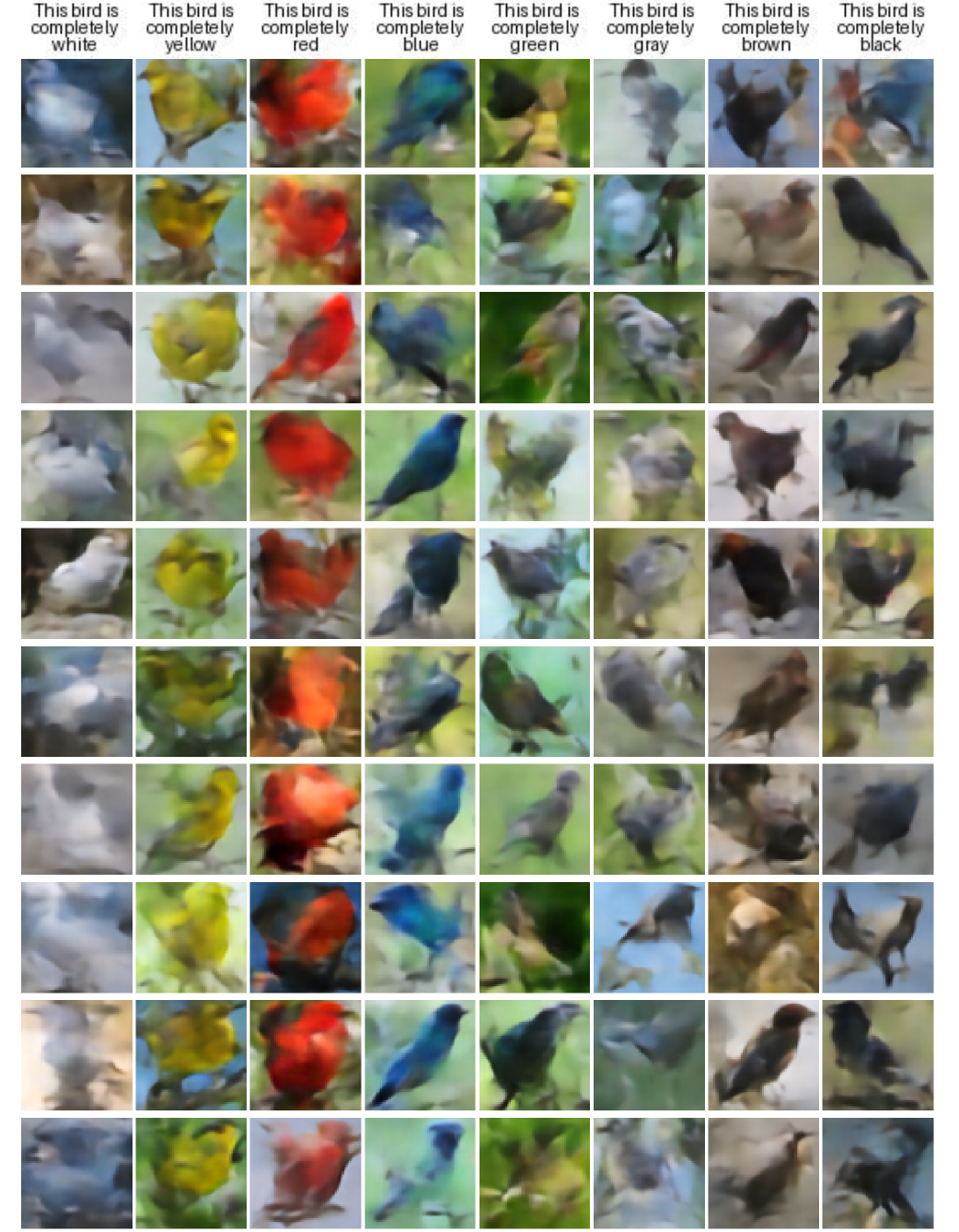}
        \caption{ MMVAE+}
    \end{subfigure}

    \vspace{0.5em} 

    \begin{subfigure}{0.32\textwidth}
        \includegraphics[width=\linewidth]{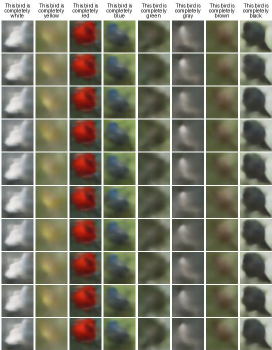}
        \caption{ MWBVAE}
    \end{subfigure}
    \hfill
    \begin{subfigure}{0.32\textwidth}
        \includegraphics[width=\linewidth]{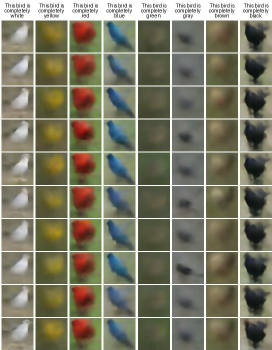}
        \caption{ CoDEVAE}
    \end{subfigure}
    \hfill
    \begin{subfigure}{0.32\textwidth}
        \includegraphics[width=\linewidth]{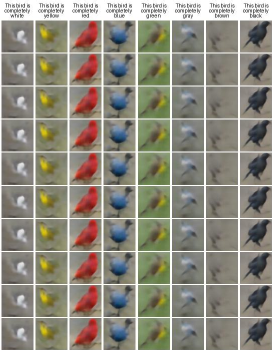}
        \caption{ HELVAE}
    \end{subfigure}
    \caption{Qualitative results of caption-to-image generation on the CUB Image-Captions dataset, used to assess generative coherence across models, with each model evaluated at its best $\beta \in \{1, 2.5, 5\}$.}
    \label{fig:cub_color}
\end{figure}

\begin{figure}[h]
    \centering
    \begin{subfigure}{0.32\textwidth}
        \includegraphics[width=\linewidth]{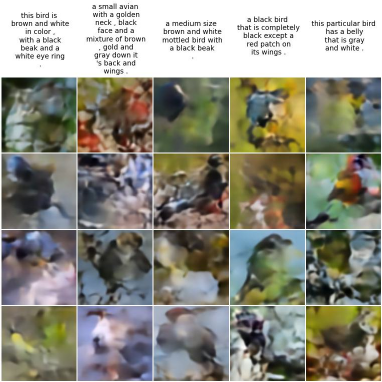}
        \caption{ MVAE}
    \end{subfigure}
    \hfill
    \begin{subfigure}{0.32\textwidth}
        \includegraphics[width=\linewidth]{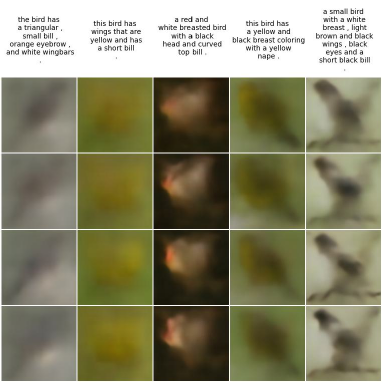}
        \caption{ MMVAE}
    \end{subfigure}
    \hfill
    \begin{subfigure}{0.32\textwidth}
        \includegraphics[width=0.973\linewidth]{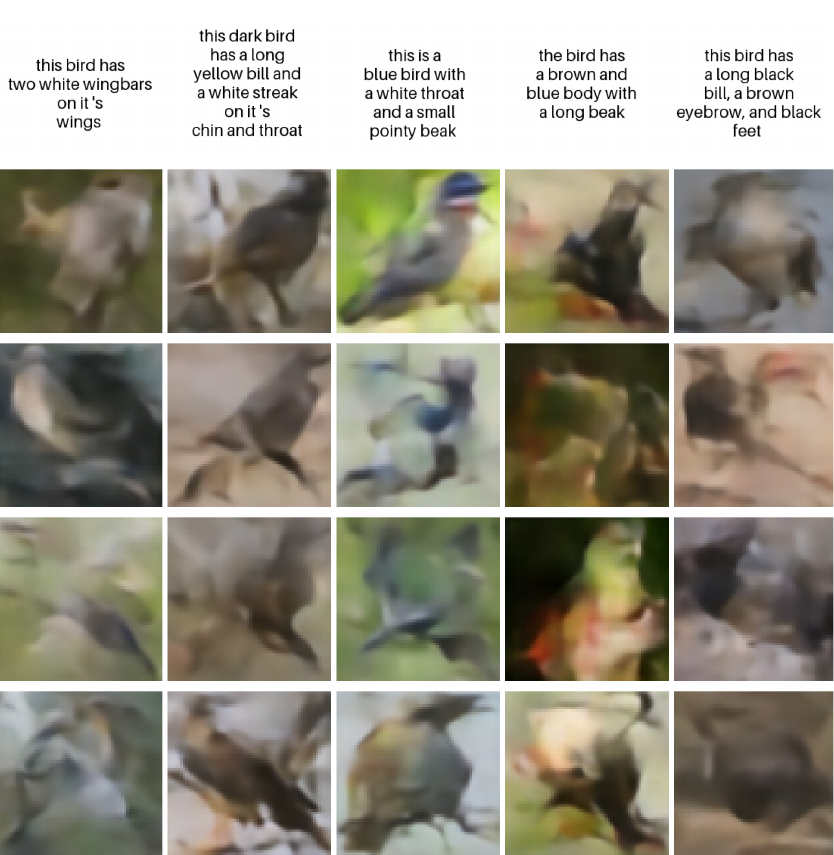}
        \caption{ MMVAE+}
    \end{subfigure}

    \vspace{0.5em} 

    \begin{subfigure}{0.32\textwidth}
        \includegraphics[width=\linewidth]{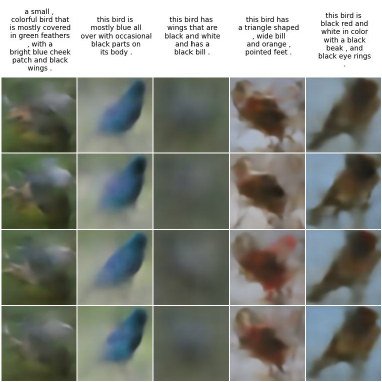}
        \caption{ MWBVAE}
    \end{subfigure}
    \hfill
    \begin{subfigure}{0.32\textwidth}
        \includegraphics[width=\linewidth]{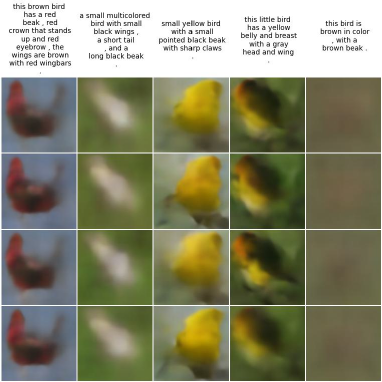}
        \caption{ CoDEVAE}
    \end{subfigure}
    \hfill
    \begin{subfigure}{0.32\textwidth}
        \includegraphics[width=\linewidth]{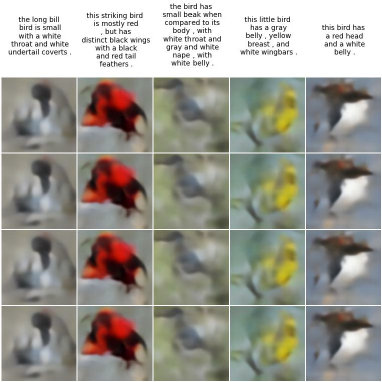}
        \caption{ HELVAE}
    \end{subfigure}
    \caption{ Qualitative results of caption-to-image generation for the CUB Image-Captions dataset, with each model evaluated at its best $\beta \in \{1, 2.5, 5\}$.}
    \label{fig:cub_text_to_img}
\end{figure}

Figure~\ref{fig:cub_color} shows the generated samples used to assess generative coherence across models on this dataset. The qualitative results of HELVAE are consistent with the quantitative results: the generated images are coherent with the captions and exhibit high visual quality. Figure~\ref{fig:cub_text_to_img} shows caption-to-image generations. For MVAE, which is a product-based approach, we observe substantial variation in the generated images but low coherence. MMVAE, MWBVAE, and CoDEVAE often collapse modality-specific features to average values, yielding blurred images. In contrast, HELVAE and MMVAE+ avoids this collapse, producing images that are both high in quality and coherent with the captions. Moreover, in Figure~\ref{fig:cub_text_to_img}, HELVAE generates samples where colors align well with the captions, and the faces and beaks are more structured, yielding coherent bird shapes.

\subsection{Additional results: Bimodal CelebA}
\label{app:celeba}

\begin{figure}[h]
\begin{center}
\includegraphics[width=\textwidth]{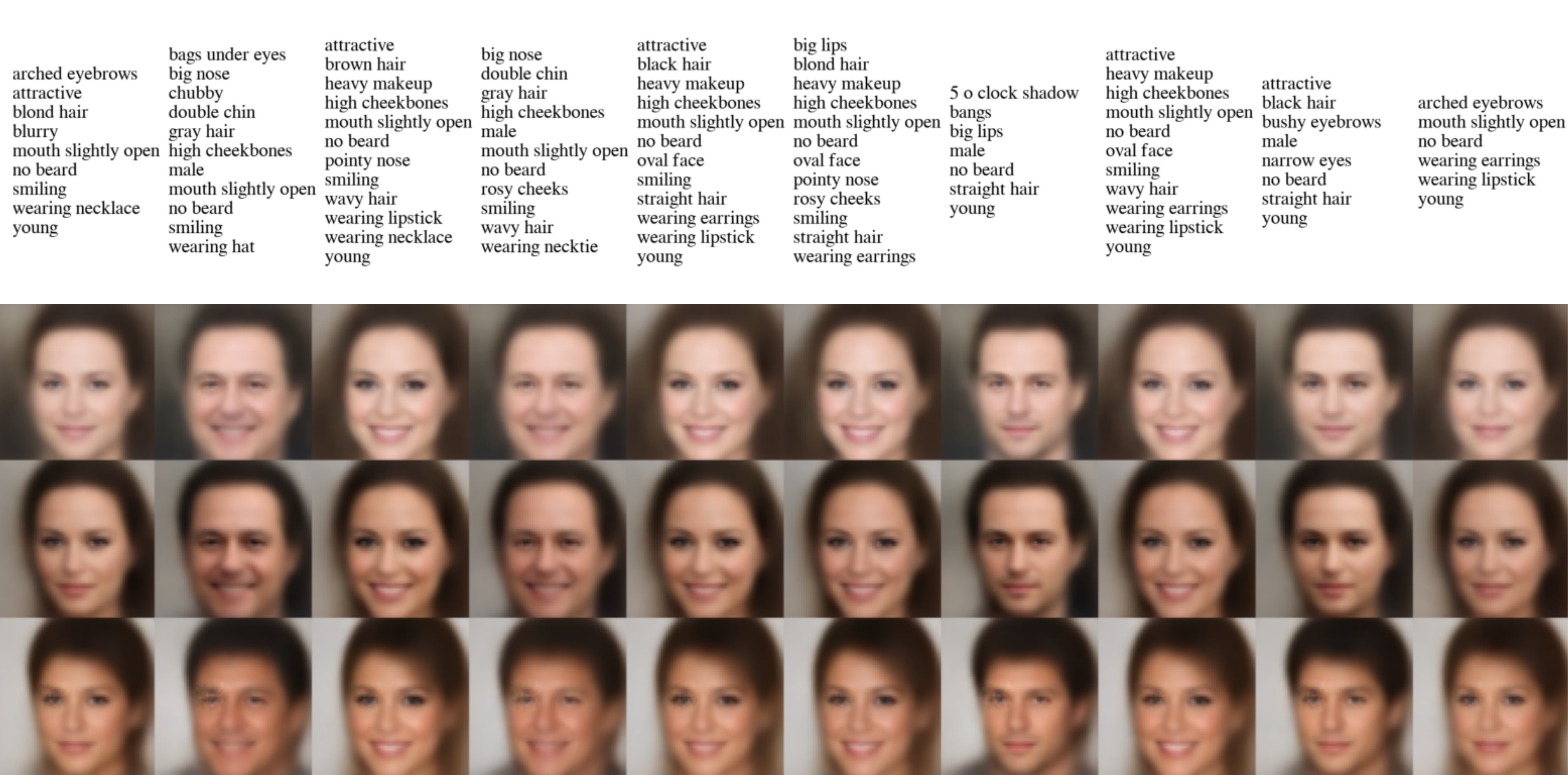}
\end{center}
\caption{ Qualitative results on the bimodal CelebA dataset using MoHELVAE, with $\beta = 1$. Each column shows images conditionally generated from the text shown at the top.}
\label{fig:celeba_text_to_img}
\end{figure}

\begin{figure}[h]
\begin{center}
\includegraphics[width=0.69\textwidth]{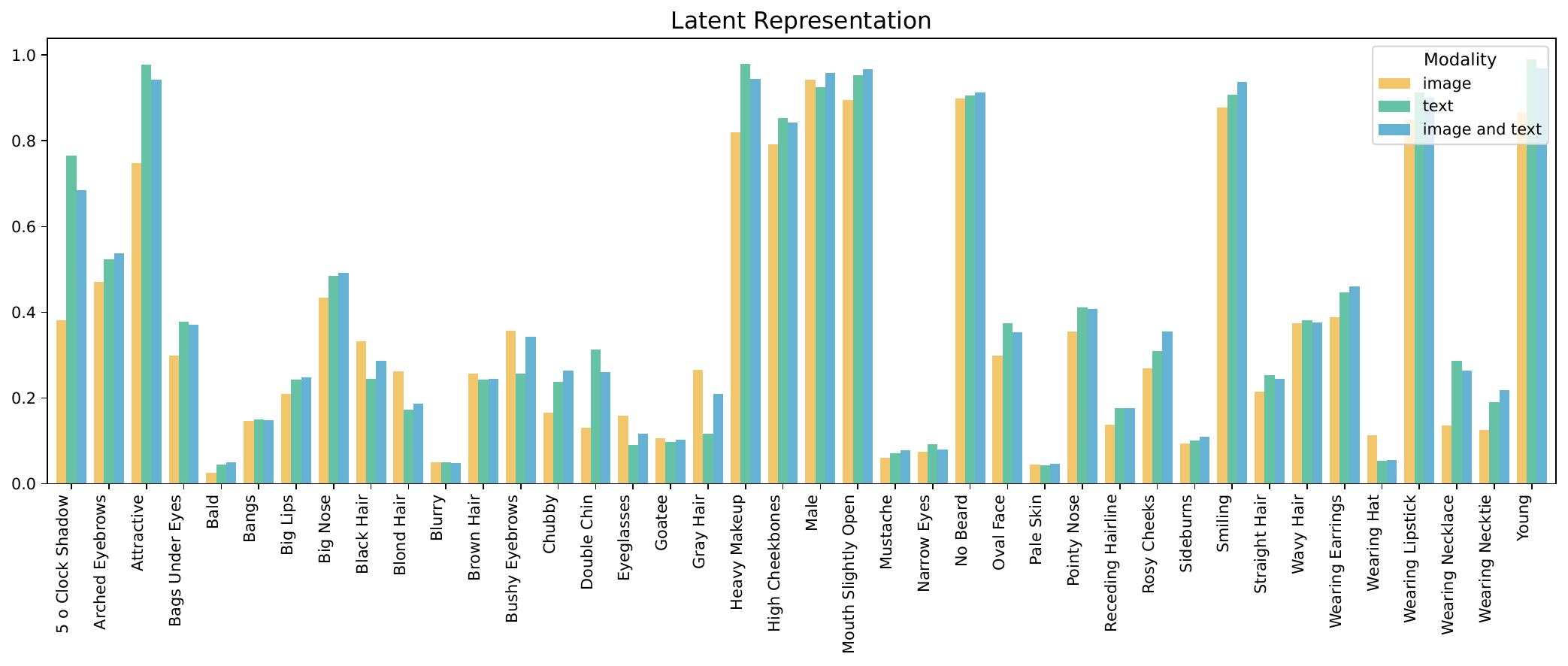}
\end{center}
\vspace{-0.1in}
\caption{ Latent representation ($\uparrow$) of the bimodal CelebA dataset using the HELVAE model, with $\beta = 1$.}
\vspace{-0.1in}
\label{fig:celeba_lr}
\end{figure}

\begin{figure}[h]
    \begin{subfigure}{\textwidth}
        \centering
        \includegraphics[width=0.69\linewidth]{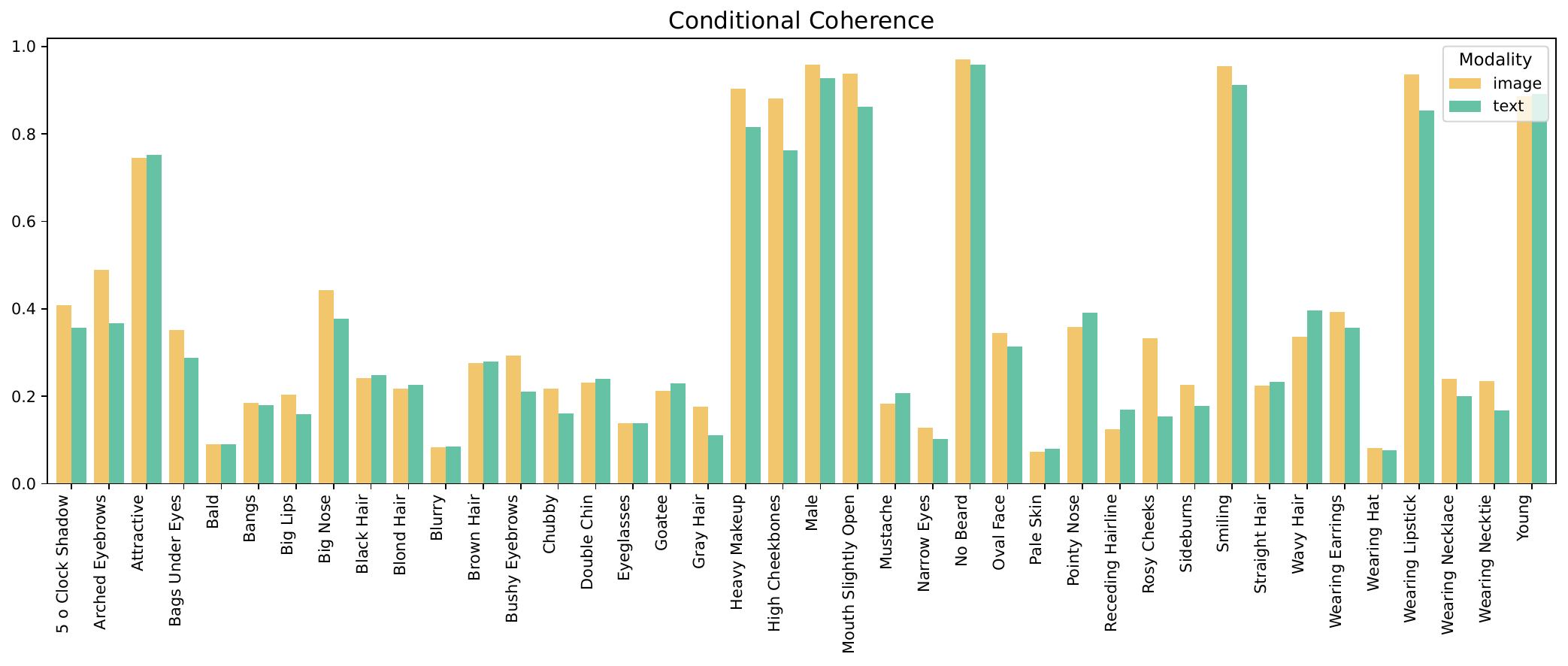}
        \caption{ Input modality: image}
    \end{subfigure}
    \hfill
    \begin{subfigure}{\textwidth}
        \centering
        \includegraphics[width=0.69\linewidth]{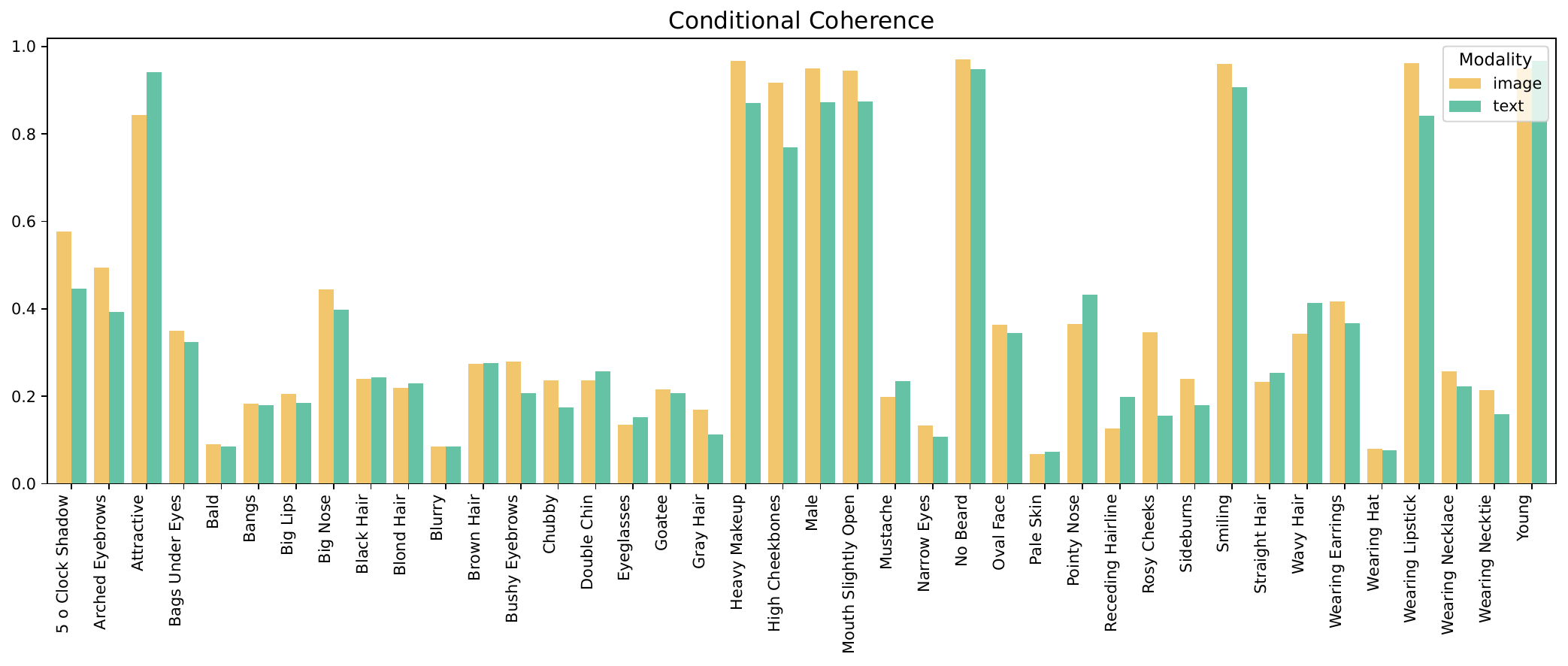}
        \caption{ Input modality: text}
    \end{subfigure}
    \hfill
    \begin{subfigure}{\textwidth}
        \centering
        \includegraphics[width=0.69\linewidth]{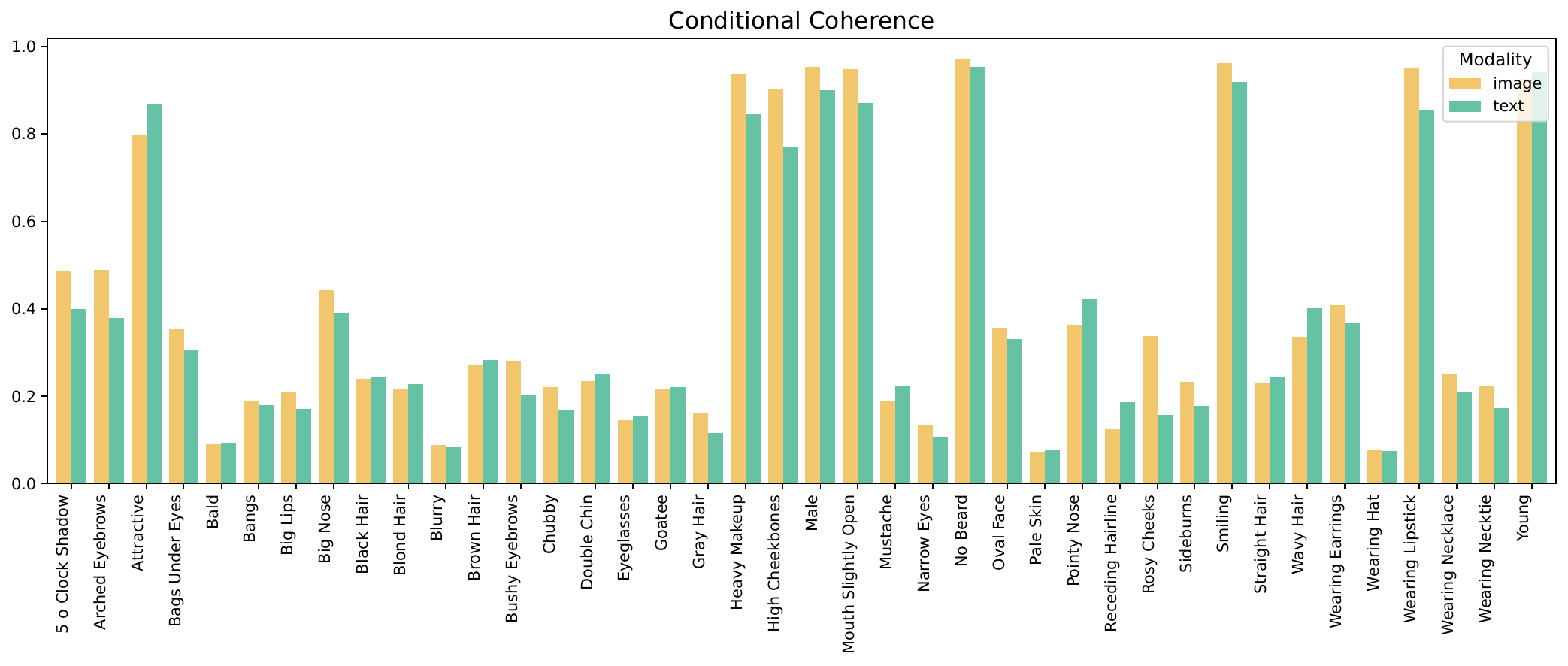}
        \caption{ Input modality: image and text}
    \end{subfigure}
    \vspace{-0.1in}
    \caption{ Conditional coherence ($\uparrow$) on the bimodal CelebA dataset using the HELVAE model, with $\beta = 1$. In each subplot, images and texts are generated conditionally from the modality or subset of modalities indicated in the caption.}
    \vspace{-0.1in}
    \label{fig:celeba_coh}
\end{figure}

We show the distribution of evaluations for each attribute in Figure~\ref{fig:celeba_lr} (latent representation) and Figure~\ref{fig:celeba_coh} (generation coherence). HELVAE achieves good performance on most large attributes, though smaller ones remain more challenging. Across all input modalities, latent representation quality and generative coherence remain consistent, indicating that the model can generate effectively even when a modality is missing. Figure~\ref{fig:celeba_text_to_img} shows that global attributes such as gender and smiling are captured well, consistent with the latent representations.

\begin{table}[h]
\caption{ Generative quality and generative coherence on the PolyMNIST dataset. Each table shows the results for the compared models for a different $\beta$ values, and rankings for each metric are reported accordingly.}
\vspace{-0.1in}
\label{tab:pm_all_results}
\begin{center}
\begin{tabular}{rcccc}
& \multicolumn{2}{c}{Conditional} & \multicolumn{2}{c}{Unconditional} \\
$\beta=1$ & Coherence ($\uparrow$) & FID ($\downarrow$) & Coherence ($\uparrow$) & FID ($\downarrow$) \\
\midrule
MVAE       & 0.176 {\tiny $\pm$ 0.007 (7)} &  \textbf{52.58 {\tiny $\pm$ 0.42 (1)}} & 0.000 {\tiny $\pm$ 0.000 (7)} &  \textbf{51.08 {\tiny $\pm$ 0.33 (1)}} \\
MMVAE      & 0.793 {\tiny $\pm$ 0.015 (5)} & 219.21 {\tiny $\pm$ 3.38 (7)} & \textbf{0.197 {\tiny $\pm$ 0.004 (1)}} & 165.66 {\tiny $\pm$ 4.42 (7)} \\
MoPoE      & 0.774 {\tiny $\pm$ 0.023 (6)} & 203.92 {\tiny $\pm$ 4.58 (5)} & 0.019 {\tiny $\pm$ 0.002 (6)} & 107.69 {\tiny $\pm$ 1.51 (4)} \\
MMVAE+     & 0.796 {\tiny $\pm$ 0.012 (4)} & 118.89 {\tiny $\pm$ 5.40 (3)} & 0.155 {\tiny $\pm$ 0.011 (2)} & 117.97 {\tiny $\pm$ 4.05 (6)} \\
MWBVAE     & 0.850 {\tiny $\pm$ 0.029 (2)} & 206.88 {\tiny $\pm$ 4.86 (6)} & 0.046 {\tiny $\pm$ 0.002 (4)} & 111.45 {\tiny $\pm$ 4.79 (5)} \\
CoDEVAE    & 0.817 {\tiny $\pm$ 0.010 (3)} & 196.64 {\tiny $\pm$ 3.63 (4)} & 0.025 {\tiny $\pm$ 0.001 (5)} & 102.61 {\tiny $\pm$ 2.41 (2)} \\
\textbf{HELVAE}    & \textbf{0.910 {\tiny $\pm$ 0.001 (1)}} & 116.83 {\tiny $\pm$ 0.62 (2)} & 0.142 {\tiny $\pm$ 0.011 (3)} & 106.05 {\tiny $\pm$ 0.25 (3)} \\
\\
& \multicolumn{2}{c}{Conditional} & \multicolumn{2}{c}{Unconditional} \\
$\beta=2.5$ & Coherence ($\uparrow$) & FID ($\downarrow$) & Coherence ($\uparrow$) & FID ($\downarrow$) \\
\midrule
MVAE       & 0.475 {\tiny $\pm$ 0.053 (7)} &  \textbf{64.01 {\tiny $\pm$ 1.45 (1)}} & 0.019 {\tiny $\pm$ 0.004 (7)} &  \textbf{63.87 {\tiny $\pm$ 1.63 (1)}} \\
MMVAE      & 0.798 {\tiny $\pm$ 0.008 (6)} & 218.62 {\tiny $\pm$ 1.72 (7)} & 0.199 {\tiny $\pm$ 0.011 (3)} & 165.71 {\tiny $\pm$ 1.83 (7)} \\
MoPoE      & 0.802 {\tiny $\pm$ 0.020 (5)} & 208.48 {\tiny $\pm$ 1.66 (5)} & 0.113 {\tiny $\pm$ 0.026 (5)} & 116.34 {\tiny $\pm$ 1.28 (5)} \\
MMVAE+     & 0.875 {\tiny $\pm$ 0.018 (2)} & 115.67 {\tiny $\pm$ 11.83 (2)} & \textbf{0.408 {\tiny $\pm$ 0.016 (1)}} & 115.13 {\tiny $\pm$ 9.78 (4)} \\
MWBVAE     & 0.832 {\tiny $\pm$ 0.008 (3)} & 209.51 {\tiny $\pm$ 2.20 (6)} & 0.145 {\tiny $\pm$ 0.020 (4)} & 118.61 {\tiny $\pm$ 4.46 (6)} \\
CoDEVAE    & 0.828 {\tiny $\pm$ 0.011 (4)} & 202.81 {\tiny $\pm$ 1.86 (4)} & 0.104 {\tiny $\pm$ 0.012 (6)} & 113.20 {\tiny $\pm$ 2.61 (3)} \\
\textbf{HELVAE}     & \textbf{0.900 {\tiny $\pm$ 0.004 (1)}} & 121.78 {\tiny $\pm$ 3.21 (3)} & 0.299 {\tiny $\pm$ 0.017 (2)} & 108.21 {\tiny $\pm$ 2.44 (2)} \\
\\
& \multicolumn{2}{c}{Conditional} & \multicolumn{2}{c}{Unconditional} \\
$\beta=5$& Coherence ($\uparrow$) & FID ($\downarrow$) & Coherence ($\uparrow$) & FID ($\downarrow$) \\
\midrule
MVAE       & 0.410 {\tiny $\pm$ 0.033 (7)} &  \textbf{96.56 {\tiny $\pm$ 0.67 (1)}} & 0.019 {\tiny $\pm$ 0.057 (7)} &  \textbf{96.75 {\tiny $\pm$ 0.73 (1)}} \\
MMVAE      & 0.792 {\tiny $\pm$ 0.011 (5)} & 218.32 {\tiny $\pm$ 0.77 (7)} & 0.197 {\tiny $\pm$ 0.014 (6)} & 166.00 {\tiny $\pm$ 1.50 (7)} \\
MoPoE      & 0.812 {\tiny $\pm$ 0.010 (4)} & 206.79 {\tiny $\pm$ 1.59 (5)} & 0.314 {\tiny $\pm$ 0.011 (3)} & 126.27 {\tiny $\pm$ 2.67 (5)} \\
MMVAE+     & 0.835 {\tiny $\pm$ 0.063 (2)} & 121.68 {\tiny $\pm$ 13.01 (2)} & \textbf{0.452 {\tiny $\pm$ 0.095 (1)}} & 120.52 {\tiny $\pm$ 11.16 (3)} \\
MWBVAE     & 0.790 {\tiny $\pm$ 0.022 (6)} & 211.23 {\tiny $\pm$ 2.37 (6)} & 0.248 {\tiny $\pm$ 0.012 (5)} & 131.39 {\tiny $\pm$ 3.58 (6)} \\
CoDEVAE    & 0.826 {\tiny $\pm$ 0.010 (3)} & 206.04 {\tiny $\pm$ 1.90 (4)} & 0.302 {\tiny $\pm$ 0.032 (4)} & 125.63 {\tiny $\pm$ 1.60 (4)} \\
\textbf{HELVAE}     & \textbf{0.887 {\tiny $\pm$ 0.006 (1)}} & 132.35 {\tiny $\pm$ 0.92 (3)} & 0.399 {\tiny $\pm$ 0.006 (2)} & 116.27 {\tiny $\pm$ 1.19 (2)} \\
\\
& \multicolumn{2}{c}{Conditional} & \multicolumn{2}{c}{Unconditional} \\
$\beta=10$ & Coherence ($\uparrow$) & FID ($\downarrow$) & Coherence ($\uparrow$) & FID ($\downarrow$) \\
\midrule
MVAE        & 0.372 {\tiny $\pm$ 0.006 (7)} & \textbf{137.33 {\tiny $\pm$ 0.06 (1)}} & 0.056 {\tiny $\pm$ 0.010 (7)} & 137.67 {\tiny $\pm$ 0.41 (2)} \\
MMVAE      & 0.803 {\tiny $\pm$ 0.011 (2)} & 217.84 {\tiny $\pm$ 3.90 (7)} & 0.185 {\tiny $\pm$ 0.013 (6)} & 164.66 {\tiny $\pm$ 2.62 (7)} \\
MoPoE      & 0.726 {\tiny $\pm$ 0.009 (4)} & 197.64 {\tiny $\pm$ 1.57 (4)} & 0.397 {\tiny $\pm$ 0.017 (4)} & 147.95 {\tiny $\pm$ 1.35 (4)} \\
MMVAE+     & 0.651 {\tiny $\pm$ 0.138 (6)} & 168.08 {\tiny $\pm$ 29.47 (3)} & 0.401 {\tiny $\pm$ 0.143 (3)} & 162.94 {\tiny $\pm$ 27.75 (6)} \\
MWBVAE     & 0.686 {\tiny $\pm$ 0.008 (5)} & 211.89 {\tiny $\pm$ 1.51 (6)} & 0.297 {\tiny $\pm$ 0.002 (5)} & 152.39 {\tiny $\pm$ 1.50 (5)} \\
CoDEVAE    & 0.771 {\tiny $\pm$ 0.001 (3)} & 197.76 {\tiny $\pm$ 0.93 (5)} & 0.431 {\tiny $\pm$ 0.006 (2)} & 144.99 {\tiny $\pm$ 1.32 (3)} \\
\textbf{HELVAE}   & \textbf{0.852 {\tiny $\pm$ 0.008 (1)}} & 154.63 {\tiny $\pm$ 0.73 (2)} & \textbf{0.508 {\tiny $\pm$ 0.019 (1)}} & \textbf{134.54 {\tiny $\pm$ 1.10 (1)}} \\
\end{tabular}
\end{center}
\end{table}

\end{document}

%% file: math_commands.tex
\usepackage{amsmath,amsfonts,bm}

\def\eqref#1{equation~\ref{#1}}

\def\1{\bm{1}}

\def\vmu{{\bm{\mu}}}
\def\vsigma{{\bm{\sigma}}}

\def\vx{{\bm{x}}}

\def\vz{{\bm{z}}}

\def\mI{{\bm{I}}}

\def\mM{{\bm{M}}}

\def\mV{{\bm{V}}}

\def\mSigma{{\bm{\Sigma}}}

\DeclareMathAlphabet{\mathsfit}{\encodingdefault}{\sfdefault}{m}{sl}
\SetMathAlphabet{\mathsfit}{bold}{\encodingdefault}{\sfdefault}{bx}{n}

\def\sR{{\mathbb{R}}}

\def\sX{{\mathbb{X}}}

\newcommand{\KL}{\mathrm{KL}}

\newcommand{\normltwo}{L^2}

\DeclareMathOperator*{\argmin}{arg\,min}